%% file: main.tex
\documentclass{article}

\usepackage{packages}
\usepackage{PRIMEarxiv}

\title{Deep Learning and Linear Programming for Automated Ensemble Forecasting and Interpretation}

\author{
  Lars Lien Ankile \& Kjartan Krange \\
  Department of Industrial Economics \\
  Norwegian University of Science and Technology \\
  Trondheim, Norway\\
  \texttt{\{larslank,kjartkra\}@stud.ntnu.no} \\
}

\begin{document}

\maketitle

\import{./Sections}{0_Abstract}

\keywords{forecasting \and time-series \and machine learning \and long short-term memory network \and autoencoder \and ensemble model}


\import{./Sections}{1_Introduction}

\import{./Sections}{2_Theory}
\import{./Sections}{3_Methods}
\import{./Sections}{4_Results}
\import{./Sections}{5_Discussion}
\import{./Sections}{6_Conclusion}

\newpage

\bibliographystyle{unsrt}
\bibliography{references}

\newpage

\appendix

\import{./Appendices}{Theory}

\import{./Appendices}{Methods}

\import{./Appendices}{Results}

\end{document}

%% file: Sections/0_Abstract.tex
\begin{abstract}
This paper presents an ensemble forecasting method that shows strong results on the M4 Competition dataset by decreasing feature and model selection assumptions, termed DONUT (DO Not UTilize human beliefs). Our assumption reductions, primarily consisting of auto-generated features and a more diverse model pool for the ensemble, significantly outperform the statistical, feature-based ensemble method FFORMA by \cite{montero2020fforma}.

We also investigate feature extraction with a Long Short-term Memory Network (LSTM) Autoencoder and find that such features contain crucial information not captured by standard statistical feature approaches. The ensemble weighting model uses LSTM and statistical features to combine the models accurately. The analysis of feature importance and interaction shows a slight superiority for LSTM features over the statistical ones alone. Clustering analysis shows that essential LSTM features differ from most statistical features and each other. We also find that increasing the solution space of the weighting model by augmenting the ensemble with new models is something the weighting model learns to use, thus explaining part of the accuracy gains.

Moreover, we present a formal ex-post-facto analysis of an optimal combination and selection for ensembles, quantifying differences through linear optimization on the M4 dataset. 

Our findings indicate that classical statistical time series features, such as trend and seasonality, alone do not capture all relevant information for forecasting a time series. On the contrary, our novel LSTM features contain significantly more predictive power than the statistical ones alone, but combining the two feature sets proved the best in practice.

\end{abstract}

%% file: Sections/1_Introduction.tex
\section{Introduction}

This paper investigates ensemble methods for forecasting, meaning the combination of simple method forecasts to improve accuracy and generalizability. We present the DO Not Utilize human assumptions (DONUT) method for ensemble combination forecasting. Our main contribution is to reduce assumptions for forecasting using time series features created from a Long Short-Term.
Preprint submitted to International Journal of Forecasting	February 15, 2022
Memory (LSTM) Neural Network (NN) as input to a weighting of 14 statistical forecasting models. We also present an ex-post-facto analysis on optimal ensembles, and provide a theoretical framework for ensembles of different flavors.
This work has two central hypotheses:

\textbf{Hypothesis 1: Fewer assumptions leads to better results.}

It is possible to improve forecasting approaches by reducing human time series assumptions (defined as model design and implementation choices not based on a particular quantitative finding, rather human ``judgement'') in the feature selection and the weighting of the ensemble. In addition, large sets of data make learning and fitting hard forecasting problems possible.

\textbf{Hypothesis 2: Machine learning methods can find relationships that humans are not able to uncover.}

Machine learning methods can find features that contain richly descriptive qualities of a time series that are not necessarily linear, which traditional statistical measures cannot. Thus, sets of statistical and machine-learned features can complement each other. Particularly, we think they can capture complex non-linear functions of explanatory variables. 

Less centrally to this paper, but still of academic interest, we hypothesize that different series exert different levels of forecastability. In particular, we hypothesize that Demographic, Industry, and Macro series are more easily forecast than Finance and Micro, here termed the hypothesis of type. Furthermore, we believe ex-ante that low-frequency data is more easily forecast than high-frequency, since low-frequency data aggregate over more data, and therefore could be less susceptible to noise, here termed the hypothesis of sampling frequency.

In addition to exploring the hypotheses stated above, we have a two-fold goal in this paper: 1) to interpret the results from the best-performing models, and understand what features of the time-series and models are essential for creating robust ensembles and accurate forecasts, thereby separating reasonable assumptions from unproductive ones; and 2) to improve upon the results in the M4 Competition by augmenting the methods and approaches that achieved the best results.

We start with a brief primer on the development of forecasting. The first studies conducted to assess the quality and accuracy of forecasting methods were conducted in the early 70s by Reid, Newbold, and Granger \cite{makridakis2000m3, reid1969comparative, newbold1974experience}. Already in 1974, Newbold and Granger found that combining naïve forecasting methods in most cases produces better results than each of the methods individually \cite{newbold1974experience}. This finding has been supported several times, e.g., in forecasting competitions and benchmarks like M1, M2, M3, where the complex methods did not outperform the more straightforward methods \cite{makridakis1982accuracy, makridakis2000m3}. Many other researchers further corroborate these findings as well \cite{armstrong1992error, fildes1998generalising}.

The M competitions \cite{makridakis1982accuracy, makridakis2000m3, M4results} are an important global competition to improve the field of forecasting has been paramount to our research. An analysis of the methods used in the M2 Competition found that the simple average over the models in an ensemble outperformed a weighted average based on the in-sample covariance matrix of fitting errors, although this method also outperformed the individual methods \cite{makridakis1982accuracy}. This finding would indicate that when creating weighted averages, one must avoid over-fitting to the in-sample. Nevertheless, in the fourth Makridakis Competition, M4, the top six methods significantly outperformed the simple ensemble benchmarks, albeit at a much higher computational cost \cite{makridakis2020predicting}, hence showing that the field has progressed. All of these methods with the exception of one use ensembles. Moreover, the top two methods utilize modern machine learning (ML) techniques to some degree, an uplifting observation, as it shows that there are advances to be made within forecasting, and that with better methods and more computing power, people can be better able to model the future, and plan and make vital decisions accordingly.
Although they exist the M5- and M6 Competition, are less relevant to us for various reasons.  

As far as we know our model is unique in using novel LSTM-features combined with classical ones to in a model combination manner forecast time series. Our framework for forecasting using model combination ensembles, termed the \textit{DONUT Approach}, short for DO Not Utilize human assumptions, aims to reduce the amount of human judgment needed about the nature of the problem and data at hand while producing accurate forecasts. 

In this paper we start by presenting \autoref{sec:theory} where some relevant theory underlying our model and related papers on which we draw is discussed. Next, in \autoref{sec:methods}, we briefly present some of the methods we use, as well as an analysis of the dynamics of optimal ensembles. Then, in \autoref{sec:results}, we present the most compelling results from the DONUT model, including analyses of the performance compared to \cite{montero2020fforma}, and an analysis of the features created by the autoencoder. Finally, \autoref{sec:discussion}, we reflect upon the results and some implications before we shed light upon where we believe we can find further performance enhancements.

%% file: Sections/2_Theory.tex
\section{Literature on Forecasting}
\label{sec:theory}

This section presents some theory and related papers laying the foundation for this work. First, we discuss the forecasting problem and loss functions in the M Competition, from which we draw many ideas and data. Next, we discuss the usage of LSTMs in forecasting and some theoretical work on ensemble forecasting. Lastly, we describe the models we have added to our final ensemble and theory on Neural Networks, of which our weighting model is an example.

\subsection{Forecasting Problem Definition and Accuracy Measurement}
\label{sec:loss_functions}

In this paper, we work with the problem of univariate point forecasts for time series (TS) in discrete time. The problem of univariate forecasting is as follows: Given a forecast horizon $h$, and a set of observed previous values of the TS  $\mathbf{y}_t=[y_1, ..., y_t] \in \mathbb{R}^t$ of length $t$, produce a vector of forecasts of length $h$ $\mathbf{\Hat{y}}_{t+h}=[\Hat{y}_{t+1}, ..., \Hat{y}_{t+h}] \in \mathbb{R}^h$ that is ``similar'' to $\mathbf{y}_{t+h}=[y_{t+1}, ..., y_{t+h}] \in \mathbb{R}^h$. A forecasting model has the form

\begin{align}
    f(\mathbf{y}_t; \mathbf{\theta}, \mathcal{M}) = \mathbf{\Hat{y}}_{t+h},
\end{align}

where $\theta$ is the model parameters and $\mathcal{M}$ is the set of available models in the ensemble model, described in detail in \autoref{sec:theory-ensemble-forecasting}.

To assess the performance of a model, one needs a way to quantify the ``similar'' property, which is a function from $\mathbf{\Hat{y}_{t}}$ mapped to how well or badly one does on unseen data. If this function is defined neatly, one can optimize it by tweaking the parameters $\theta$ of $f$. The objective function is the function we want to minimize or maximize, and is often termed \textit{loss function} in machine learning literature.

We have opted to use the same loss functions employed in the M4 Competition during the fitting and testing of our model: Overall OWA, sMAPE, and MASE, which are defined in \cite{M4results}. First and foremost, this allows us to compare our performance to the performance of the models in the M4 Competition. Moreover, the loss functions have the property of being continuous, as mentioned as being necessary above. Lastly, the measures allow for comparing across TS of different levels, unlike, e.g., the mean absolute error \cite{hyndman2018forecasting}. The primary measure is the OWA loss function, which is a normalized combination of sMAPE and MASE, two standard accuracy measures used in the field of forecasting \cite{petropoulos2020forecasting}.

Note that, through some algebraic manipulation, one can rewrite the MASE loss as the objective function of a linear optimization program. Therefore MASE, and not OWA, will be used in a posteriori optimzation.

\subsection{M4 Competition and Data}

This paper presents a model based on an extensive analysis of the findings of the Makridakis M4 forecasting Competition \cite{M4results}. We have combined ideas, primarily from the two winning papers by \cite{smyl2020hybrid} and \cite{montero2020fforma}, as well as augmenting the methods with our ideas and other methods from the literature. The M4 Competition hosted 248 contestants from 46 countries, of whom 49 teams provided valid forecasts for the 100,000 times series by \cite{M4results}.

 The FFORMA model of \cite{montero2020fforma} exploits the established finding that ensembles tend to outperform their constituent parts. Furthermore, the winning paper by \cite{smyl2020hybrid} and its hybrid solution discuss that using all the available information about a time series (e.g., time-series type) is essential for accurate forecasts. Both these findings are utilized in our approach.

This paper has made an ensemble forecasting technique by augmenting the FFORMA model in several ways. In \cite{montero2020fforma}, the main idea is to use statistical features of a time series (e.g., trend, variance, seasonality) as inputs to a model that weights the output of nine more straightforward forecasting methods. These models include the naïve forecast, ARIMA, exponential smoothing, and other established models. \cite{montero2020fforma} find significant improvement from averaging over the models compared to picking the single best model. Their OWA loss is also 14\% smaller than simple averaging across all nine models.

In the appendix we have summarized the results from each of the previous M Competitions, which shows how publicly available forecasting methods have improved over the years. A significant finding in the first three M Competitions was that more sophisticated methods generally did not outperform simple methods such as naïve and performed worse when the test series contained considerable noise. Additionally, averaging multiple methods or combining them proved more successful in the M1 and M3 competitions. Another interesting finding from the M2 Competition was that the most significant improvements in forecasting accuracy came from measuring and extrapolating the seasonality of a series.

\subsection{LSTM in forecasting}
\label{sec:lstm}

We chose to use the Long Short-term Memory network (LSTM) for feature extraction, in part due to the paper by \cite{laptev2017time} (winner of M4, Smyl, is also co-author). The paper uses an LSTM for feature extraction, thus creating condensed input that will function as features for a neural network to make predictions. Furthermore, some statistical features are used in addition to the LSTM features, possibly giving the output even more predictive power. The authors developed new methods to predict extreme events in Uber's data, e.g., a drastic reduction in traffic on Christmas Day. Compared to current univariate machine-learned models, they improve the forecast accuracy by 2\% - 18\%. They also showed decent results on the M3 Competition data, outperforming several traditional methods, hence indicating a good generalization.

\cite{laptev2017time} find that a high number, long length, and existence of a correlation between time series indicate that NNs could be superior to statistical methods. However, statistical methods still seem to be superior in more straightforward cases.
A resulting hypothesis is that using NNs to combine statistical models allows for the discovery of complex relationships given enough data and the possibility of weighting simple statistical models when the signal-to-noise ratio is low, or the number of series is small. In our case, the most important result from this paper is the use of the LSTM. While \cite{laptev2017time} also uses an LSTM for forecasting, the idea of LSTM generated features is one of our core approaches.

\subsection{Ensemble Forecasting}
\label{sec:theory-ensemble-forecasting}

On a high level, there are two ways to produce forecasts. The first, and simplest, is using a single model, while the second is combining several models into a single forecast, referred to as an ensemble in the literature \cite{montero2020fforma}. In a large body of research, ensemble methods are shown to be superior to single-model approaches, starting in 1969 with \cite{bates1969combination}, further supported by many others (e.g. \cite{clemen1989combining,makridakis2000m3,timmermann2006forecast}).

There are two main ways of producing ensemble forecasts given a set of forecasting methods, \textit{Model selection} and \textit{Model combination}, with model combination generally outperforming model selection \cite{petropoulos2020forecasting}. In this work, we focus on the combination of forecasting models, which will often be referred to as \textit{model weighting}, as in effect, we create weighted averages over the models.

The ``no-free-lunch'' theorem establishes that no algorithm can outperform all other models or the random forecast when testing on all possible data \cite{wolpert1996lack}. This theorem implies that without knowing anything about a problem, one cannot assume anything about the performance of an algorithm \cite{lemke2010meta}. Of course, there will be specific problems for which one algorithm performs better than another in practice. Consequently, \cite{lemke2010meta} show that the above assumption can be relaxed by automatically extracting features from a series that will function as a proxy for domain knowledge. \cite{Goodfellow-et-al-2016} also argue that the reason model combination works are that different models will not have residuals with perfect correlation for the test set. At the same time, model diversity alone must not be the only aspect of an accurate combination of methods---it is individual accuracy as well, according to \cite{lemke2010meta}.

Lastly, an important reason for us to choose a model combination ensemble is the interpretability it offers. We find it interesting to interpret the results and weights our weighting model offers.

\subsection{Statistical Forecasting Models}
\label{sec:new-models}

In the appendix, we explain the workings of five models we have added to the existing nine models used by \cite{montero2020fforma}, and some rationale for adding each model to the set of available models $\mathcal{M}$.

\subsection{Artificial Neural Networks}

In the appendix, we summarize some background literature on Neural Networks and how they are trained and tuned. We also refer the reader to \cite{Goodfellow-et-al-2016}.

%% file: Sections/3_Methods.tex
\section{Methods}
\label{sec:methods}

In this section, we present the methods we have used in-depth, in addition to an analysis of the dynamics of optimal ensembles. We start by presenting our data-driven ensemble algorithm top-down. We then present some forecasting theory and a model to understand how ensemble loss occurs. 

\subsection{A Data-Driven Ensemble}
In several papers, we found a high degree of modeling decisions made based on a qualitative judgement, rather than empirical and quantitative investigations. We suspect that such judgements could lead to bias and erroneous assumptions. Therefore, we choose both the features used for combining statistical models and the statistical models used in the ensemble by quantitatively assessing their effectiveness, as described in \autoref{sec:hyperparam-tuning}.

\begin{figure}
    \centering
    \includegraphics[width=0.8\textwidth]{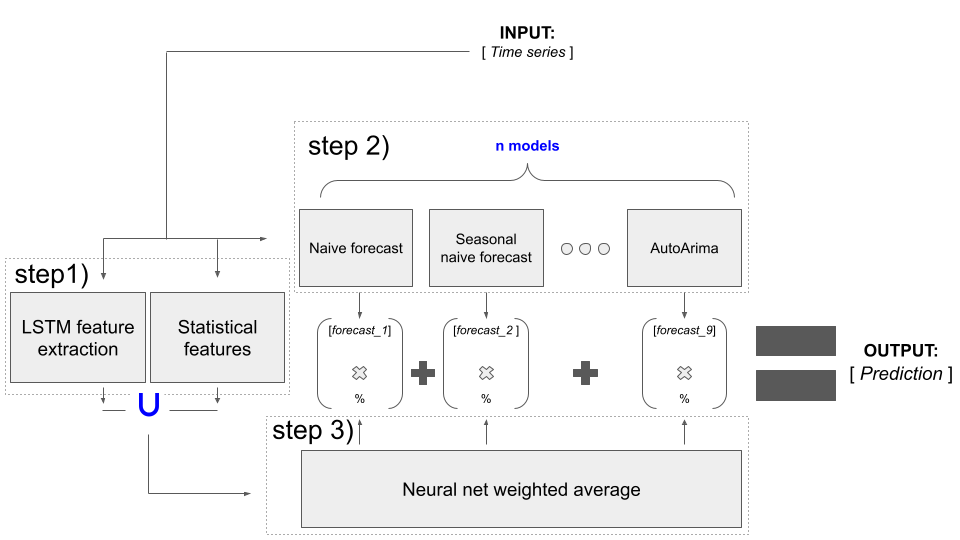}
    \caption{Model schematic. Features are generated from time series, which a weighting neural net uses to generate weighted average prediction of n=14 models. Blue marks where our model differs significantly from \cite{montero2020fforma} .}
    \label{fig:ensemble}
\end{figure}

\subsection{Algorithm}

Our algorithm consists of three primary steps, as enumerated in \autoref{fig:ensemble}. Furthermore, all our code is found in the GitHub repository: \href{https://github.com/Krankile/ensemble_forecasting}{krankile/ensemble\_forecasting}.

The following is a top-down view of how our algorithm work:

\textbf{Step 1} 
We fit a Long Short-term Memory model to the set of time series. Thereafter, we use the fitted model to create 32 novel features of the time series. We also collect statistical features based on the tsfeatures R library, as found in \cite{montero2020fforma}.

\textbf{Step 2) }
We create a forecast based on 14 models, nine of which are collected from FFORMA \cite{montero2020fforma}. The full list of all the statistical models we use are as in FFORMA, with the addition of the following: 
\begin{enumerate}
    \item OLS
    \item Ornstein Uhlenbeck Mean Reversion
    \item Local and Global Trend Model (\cite{ng2020orbit})
    \item .99 Quantile Regression
    \item .01 Quantile Regression
\end{enumerate}

\textbf{Step 3)}
We weight these forecasts based on the LSTM and statistical features, consisting of 74 features in total. We also use two meta-variables as input, the ``period'', which is the time delta of successive observations of a series, and ``type''. A weighting neural net is trained versus actual values according to OWA loss. The results are then acquired from predicting the separate test set as provided in the M4 Competition.

\subsection{Theoretical Best Forecast of Ensembles}

In this section, we shed some light on the theoretical best ensembles for a given $\mathbf{y_t}$, $\mathcal{M}$ (thus a set $\{\mathbf{\hat{y}}_{t,m}\}|m \in \mathcal{M}$) and given $\mathbf{y_{t+h}}$. $\mathbf{y^{*}_{t,M}}$ marks the theoretical absolute optimum of an ensemble $M$.

We define a linear program minimizing the absolute loss of a combination ensemble below; this is done by finding the weights which reduce loss the most. We call the optimal solution, the Implicit Ensemble Loss, or $E_{loss}$ \footnote{We note that this model implicitly finds the best weights to minimize MASE loss, this is because MASE $\propto$ MAE. We find the best absolute weights, then multiply this value by an appropriate constant to find the global minimum MASE.}. 

Given this linear program, we can dissect the possible absolute error loss in a theoretically optimal solution. One can then think of the total loss a fitted model is experiencing as the sum of a wrong weighing and the Implicit Ensemble Loss: 

\begin{align}\label{eq:T-loss}
    Total_{loss}(\mathbf{y}_t;\theta,\mathcal{M}) = P_{loss}(\mathbf{y}_t;\theta,\mathcal{M})+E_{loss}(\mathcal{M,\mathbf{y_t},\mathbf{y_{t+h}}}) \in \mathcal{L}_{\Omega}
\end{align} 

To a certain extent, we should be able to reduce the loss $P_{loss}$ from poor weighting through training, while $E_{loss}$ is something given by choice of ensemble. This formula highlights that one should take care in choosing the ensemble composition. Indeed, even with an omniscient weighting entity, the designers of an ensemble algorithm can greatly reduce its upper bound accuracy even before training. To contextualize, our model below has $P_{loss} = NN(\theta) \in \mathbf{\Hat{p}}_{\mathcal{M}}$, where $NN(\cdot)$ is an Artificial Neural Net. Although we do not know $\mathbf{y_{t+h}}$ at time $t$, the model above gives a good method for making inferences on combination ensembles. For this reason, it is only used in $t_1 \geq t+h$ to gain the knowledge of what caused a certain $T_{loss}$.  

If we run the linear program over our M4 validation set, we find the best possible loss distribution below in  \autoref{fig:opimization-14-models} and \autoref{fig:opimization-9-models} with our 14 models and the nine models as in \cite{montero2020fforma} plotted respectively. Note that most of the new models from \autoref{sec:new-models} are not given a large weight in the post facto forecast, but that they are non-negative, leading to meaning that FFORMA has a higher lower bound error. In the Implicit Ensemble Error for FFORMA, naïve is given the least weight. Furthermore, in 0\% of cases, eight models are used in the optimal ensemble made by the linear program. Although nowhere near using all 14 models, our approach still has a decreased lower bound loss than FFORMA since it uses a higher number of models. 

We find the mean ex-post-facto MASE optimum for FFORMA to be 1.077 and 0.883 for our 14 models approach on our held-out validation set, respectively.

\subsection{Ensemble Building Heuristics}

Selecting what models should be part of an ensemble is not a trivial task, and will restrict the possibility of an optimal weight of predicting. We run a greedy search in selecting a model for the ensemble by the model of all left, which results in the lowest MASE value. As such, we see how $E_{loss}(\mathcal{M})$ gradually diminishes with more models. We reason that this is because of positive correlations between already selected models and possible candidates. For instance, notice that Naive and OLS are the last models added to the ensemble. This is possibly caused by the other algorithms already holding much of their predictive power. As such, adding them to the ensemble does not predict a series more accurately.

One can also note that the greedy result has an optimal ensemble error lower than that of FFORMA already at four models. Although this is ex-post-facto, it indicates how adding some models gives expression, and that adding many statistical models provides a marginally decreasing loss reduction. One could then conclude that adding fewer models to make the weighting entities.

\begin{figure}
    \centering
    \includegraphics[width=0.9\textwidth]{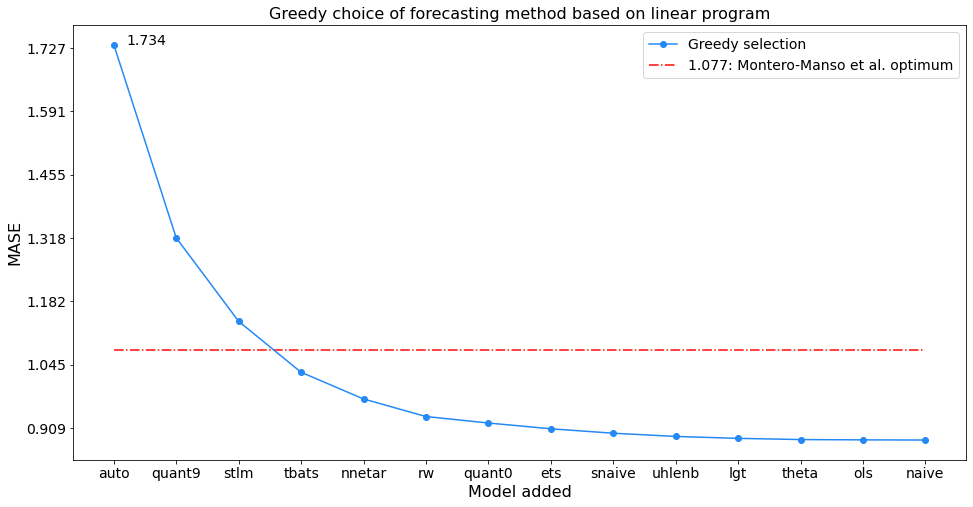}
    \caption{Greedy search for optimal ex post facto ensemble}
    \label{fig:ensemble-build}
\end{figure}

\subsection{Model selection vs. model weighting theory}

\subsubsection{Optimal Loss of Model Combinations and Selections}
\label{sec:method-proof-loss}

One can prove that the ex-post-facto model selection approach has at least as high a loss as a model weighting technique. One proof of this can be found in \autoref{proofs}.

\subsubsection{Computational proof of model weight ex-post superiority}

Using the linear program, we find that model selection ex-post-facto only amounts to 34\% of optimal solutions (\autoref{fig:opimization-9-models}) with \cite{montero2020fforma}'s ensemble, which reduces to 16\% of optimal solutions when using our 14-ensemble (\autoref{fig:opimization-9-models}). We can also calculate FFORMA's upper bound model selection score found in the table below.  

\begin{table}
    \centering\label{tab:mase-fforma}
    \footnotesize
    \singlespace
    \begin{tabular}{@{}lll@{}}
    \toprule
    \textbf{FFORMA}               & \textbf{Mean}  & \textbf{Variance} \\ \midrule
    MASE Model Combination  & 0.944 & 2.558    \\
    MASE Model selection & 1.053 & 2.707    \\ \bottomrule
    \end{tabular}
    \caption{Table of the ex post facto MASE score for optimal model selection and optimal model combination for the FFORMA ensemble of nine models.}
\end{table}

For the FFORMA ensemble, we calculate the MASE loss ex-post-facto, finding the optimal weights of the combination problem and the optimal selection problem. In line with the results in \autoref{proof:model-comb}, we see a 10.4\% reduction in MASE and a 5.5\% reduction in optimal solution variance in a combination vs. selection approach ex-post-facto. This implies that with their current ensemble, \cite{montero2020fforma} would theoretically be able to reduce their MASE by 32.1\% with selection, and 39\% in keeping their combination ensemble framework.

\subsection{Recurrent Autoencoder}
\label{sec:method-autencoder}

We use a form of Recurrent Neural Networks, called a Long Short-term Memory network (LSTM), in an Autoencoder (AE) constellation, to extract features from a time series (TS). These features will later be provided as input for our final ensemble weighting model. By using an LSTM, the aim is to create features of a TS which are not obvious to humans, or even more so, features humans cannot themselves think of. Therefore, this combines human and NN feature extraction to leverage more predictive features of a TS in forecasting.

AEs are a tool for Non-Linear Principal Component Analysis (NLPCA), and can serve similar purposes such as dimensionality reduction, visualization, and exploratory data analysis \cite{kramer1991nonlinear}. The primary advantage of using an AE is that Principal Component Analysis (PCA) can only find linear relationships between variables, while the AE places no such restrictions on the variable interactions. \cite{kramer1991nonlinear} proves that NLPCA can efficiently reduce dimensionality, and produce a lower-dimensional representation that accurately resembles the actual distribution of the underlying system parameters.

\begin{figure}
    \centering
    \includegraphics[width=0.8\textwidth]{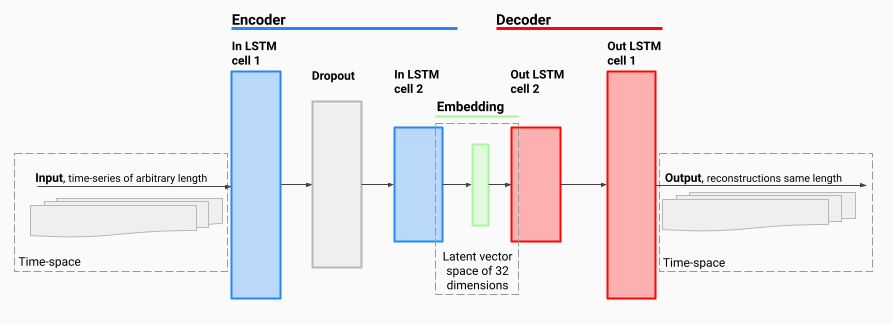}
    \caption{The LSTM model architecture. There is two LSTM layers on each side of a bottleneck. There is also a dropout layer between the two LSTMs in the encoder.}
    \label{fig:lstm-arch}
\end{figure}

A Deep Neural Net (DNN) architecture of LSTM cells was chosen to implement a time series feature extractor. The superiority of LSTMs for modeling complex temporal relationships is argued in \autoref{sec:lstm}. An Autoencoder (AE) is a DNN architecture used for unsupervised learning, which is ideal for this application as the goal is to generate features without human judgement or labeling. This is achieved by providing raw data as input to the model, and scoring the model on how well it is able to recreate the input. This procedure becomes interesting once a bottleneck is added (the green box in \autoref{fig:lstm-arch}), which makes sure that the maximal number of parameters the model can work with is limited to be strictly less than the number of input features to be recreated. This forces the Autoencoder to learn how to represent the input in a lower-dimensional space, often called a \textit{latent vector space} or \textit{vector embedding space}. Our concrete model takes TS of varying lengths, and compresses them into a time-invariant vector of length 32 that exists in a vector space in $\mathbb{R}^{32}$.

Our architecture comprises two layers of LSTM cells on each side of the bottleneck or embedding, see \autoref{fig:lstm-arch} for context. The part to the left of the blue embedding is called the Encoder, and the part to the right is the Decoder. During training, the entire system is trained to be able to replicate the input signal in the time-dimension from a compressed representation in $\mathbb{R}^{32}$. To successfully do this, the network has to learn how to extract the important features in a time series. During the inference phase, the Decoder is removed, and only the Encoder is used. This will thus produce vectors of a fixed size that represent time series of varying lengths for use in the model weighting network.

The hyperparameters used for training are summarized in \autoref{tab:lstm-hyperparam} in \autoref{app:hyperparameters}. The hyper-parameters are decided by thorough sweeps based on Bayesian search and the Hyperband pruning of runs, as described in \cite{li2017hyperband}. The searches are conducted using the online service \hyperlink{https://wandb.ai}{wandb.ai}. In short, we found that normalizing data was essential due to the problem of exploding gradients. Both min-max scaling and standardization were tested, with standardization seeming to work the best. Furthermore, enough regularization, in the form of dropout and weight decay, was also essential to prevent overfitting (see \cite{Goodfellow-et-al-2016} for an in-depth overview of regularization methods).

During training, the loss quickly fell from the initial levels to approximately one-fourth within the first 100 epochs. During the remaining 300, the loss was only reduced marginally. However, both the training and validation loss fell in lock-step, signifying that we have not reached a state of over-overfitting.

\subsection{LSTM Autoencoder Interpretation}

See \autoref{app:ae-interpretation} for results from reconstruction experiments and interpretation.

\subsubsection{Measuring Importance of Features Generated by the Autoencoder}
\label{sec:methods-feature-importance}

Neural Nets (NN) cannot be easily interpreted like, e.g., linear models. Therefore, more sophisticated ways of analyzing the relevance of variables are necessary. The technique of \textit{feature permutation}, introduced by \cite{breiman2001random}, is a widely used technique. It works by iteratively breaking the relationship between the dependent variable and each feature one-by-one by randomly scrambling that feature across the dataset while keeping everything else in place. Then, the importance of a feature is defined as the increase in the loss from the baseline when that feature is scrambled \cite{breiman2001random}. We utilize this technique to analyze the interaction between our weighting model and the autoencoder feature in \autoref{sec:results-feature-importance}. Here, the permutation importance procedure is run multiple times and subjected to $t$-ratio tests to investigate whether changes are statistically significant \cite{student1908probable}.

\subsection{Weight model}

Our weight model, as depicted by the Step 3) box in \autoref{fig:ensemble}, is a Fully Connected Neural Network. It is trained on 80/20 training data, and converged quickly to what seems to be a local minimum. In \autoref{tab:weight-model-hyperparam} in \autoref{app:hyperparameters}, the hyper-parameters of the model are presented. Its input is statistical and LSTM-generated features. The weight models outputs are 14 weights for each single statistical model in our ensemble.  

%% file: Sections/4_Results.tex
\section{Results}
\label{sec:results}

The DONUT model presented above achieves a superior result to the FFORMA model by \cite{montero2020fforma} (see \autoref{sec:results-m4-test-data} and \autoref{tab:results-m4}). Furthermore, we find that our model's predictions differ from the FFORMA model's in several interesting ways, as described in \autoref{sec:results-fforma-comparison}. Lastly, through an analysis of feature importances, we find that our model heavily utilizes the automatically extracted features of the autoencoder to produce forecasts, which is described in detail in \autoref{sec:results-feature-importance}.
Note: ``DONUT'' comprises the full approach and general model described, while ``DONUT Fanciful'' is the name of the specific trained model instance used to produce the results in the following sections.

\subsection{M4 Test Data Forecasting Results}
\label{sec:results-m4-test-data}

The final model achieves a result on the M4 test set of 0.830, 11.573, and 1.636 for OWA, sMAPE, and MASE, respectively. This result would give the ``DONUT Fanciful'' model the second place in the M4 Competition by some margin, with an OWA of 0.008 better than \cite{montero2020fforma} and 0.009 off the first-place approach by \cite{smyl2020hybrid}. In \autoref{tab:results-m4}, one can see that the second to sixth place is separated by 0.010 in OWA score. Our model surpasses the second place by almost the same amount, indicating that the DONUT approach is a significant improvement.

\begin{table}
    \centering
    \footnotesize
    \singlespace
    \begin{tabular}{@{}lllll@{}}
    \toprule
    Method & OWA & sMAPE & MASE & M4 rank \\ \midrule
    Smyl & 0.821 & 11.374 & 1.536 & 1 \\
    \textit{DONUT Fanciful} & 0.830 & 11.573 & 1.544 & 2$^\dagger$ \\
    Montero-Manso & 0.838 & 11.720 & 1.551 & 2 \\
    Pawlikowski et al. & 0.841 & 11.845 & 1.547 & 3 \\
    Jaganathan and Prakash & 0.842 & 11.695 & 1.571 & 4 \\
    Best single model (Theta) & 0.897 & 12.309 & 1.696 & 18 \\
    Simple average & 1.040 & 13.454 & 2.082 & 39 \\
    \bottomrule
    \end{tabular}
    \caption{Summary of our results compared to other relevant methods on the M4 dataset. Our method achieves a significantly better OWA result than the M4 second place, and a MASE better than both the first place and second place. $^\dagger$DONUT model did not enter competition but the rank is included to contextualize the result.}
    \label{tab:results-m4}
\end{table}

In \autoref{tab:results-test-set}, our result is broken down into the different categories and time intervals between successive observations of the M4 data set (termed type and period, respectively). The table shows that the model has a differing forecasting ability for different types of series. Most remarkably, the model achieves a statistically significant OWA of 0.384 on Monthly Demographic data at a $\alpha = 0.001$ significance level. Moreover, the Demographic type is the best overall at an OWA of 0.528. The only under-performing frequency in Demographic is the Daily, but this did not prove significant and comprises only 10 series, resulting in high a variance and low impact. It is also important to observe that the model performs better than average for Macro data at an OWA of 0.811.

Looking at the time intervals, one can observe that the model performs better than its average for Hourly, Monthly, and Quarterly data, and worse than average for Yearly, Daily, and Weekly data. In the case of the Hourly data, it seems there is a very specific type of series in the Other Category that the model responds well to. For Monthly and Quarterly, it is presumably because there is the right combination of enough data and enough time between observation for them to be less noisy than, e.g., Daily. With that in mind, it is surprising to see the Yearly frequency performing much below par.

\begin{table}
    \centering
    \footnotesize
    \singlespace
    \begin{tabular}{lrrrrrrr}
    \toprule
    {} &  Demo. &  Finance &  Industry &  Macro &  Micro &  Other &   \textbf{Mean} \\
    \midrule
    Daily     & 2.903 & 1.017$^c$ & 1.142$^c$ & 0.992$^a$ &  0.801 &  1.040$^c$ &  \textbf{0.961} \\
    Hourly    &   NaN &   NaN &   NaN &   NaN &    NaN &  0.682$^c$ &  \textbf{0.682} \\
    Monthly   & 0.384$^c$ & 0.738$^c$ & 0.721$^c$ & 0.715$^c$ &  0.752$^c$ &  0.647$^c$ &  \textbf{0.689} \\
    Quarterly & 0.671$^c$ & 0.698$^c$ & 0.622$^c$ & 0.638$^c$ &  0.659$^c$$^a$ &  0.483$^c$ &  \textbf{0.650} \\
    Weekly    & 1.398$^b$ & 0.786 & 0.863 & 1.204$^c$ &  0.830 &  1.454 &  \textbf{0.912} \\
    Yearly    & 0.996$^c$ & 1.356$^c$ & 1.420$^c$ & 1.307$^c$ &  1.205$^c$$^a$ &  1.190$^c$ &  \textbf{1.289} \\
    \midrule
    \textbf{Mean}      &        \textbf{0.528} &    \textbf{0.912} &     \textbf{0.844} &  \textbf{0.811} &  \textbf{0.854} &  \textbf{0.881} &  \textbf{0.830} \\
    \bottomrule
    \end{tabular}
    \caption{OWA loss for the method on the test set and for each subset of the data. The main finding is a OWA loss of 0.830, significantly better than \cite{montero2020fforma}. Significance levels: $^a$ $\alpha=0.05$, $^b$ $\alpha=0.01$, and $^c$ $\alpha=0.001$.}
    \label{tab:results-test-set}
\end{table}

\begin{figure}
    \centering
    \includegraphics[width=0.8\textwidth]{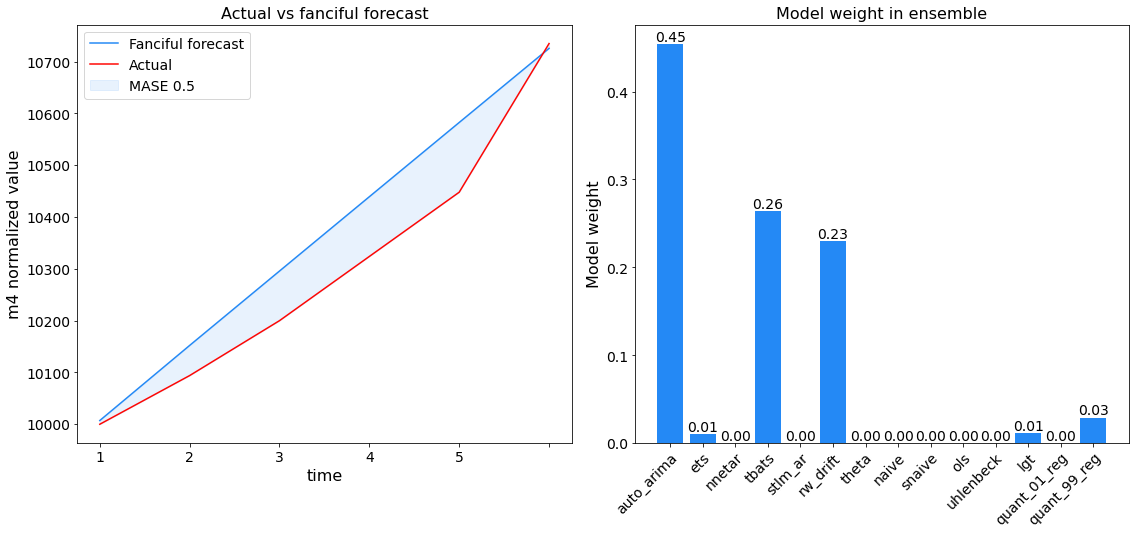}
    \caption{Left, the actual value compared to the forecast for a successful forecast. Right, the weights assigned to each model in the forecast on the left.}
    \label{fig:good-fanciful}
\end{figure}

\begin{figure}
    \centering
    \includegraphics[width=0.8\textwidth]{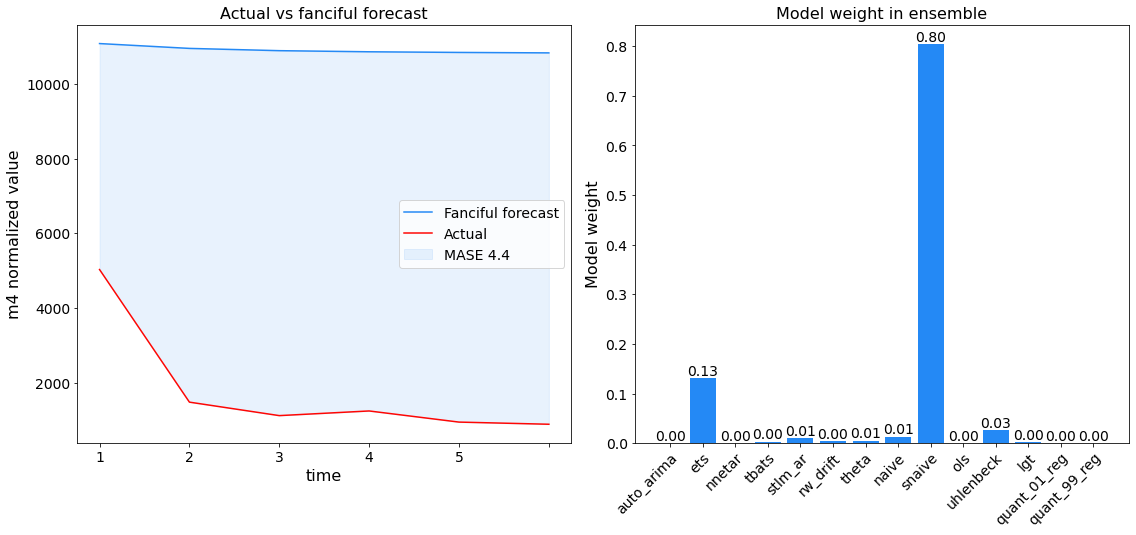}
    \caption{Left, the actual value compared to the forecast for a successful forecast. Right, the weights assigned to each model in the forecast on the left.}
    \label{fig:bad-fanciful}
\end{figure}

\begin{figure}
    \centering
    \includegraphics[width=0.8\textwidth]{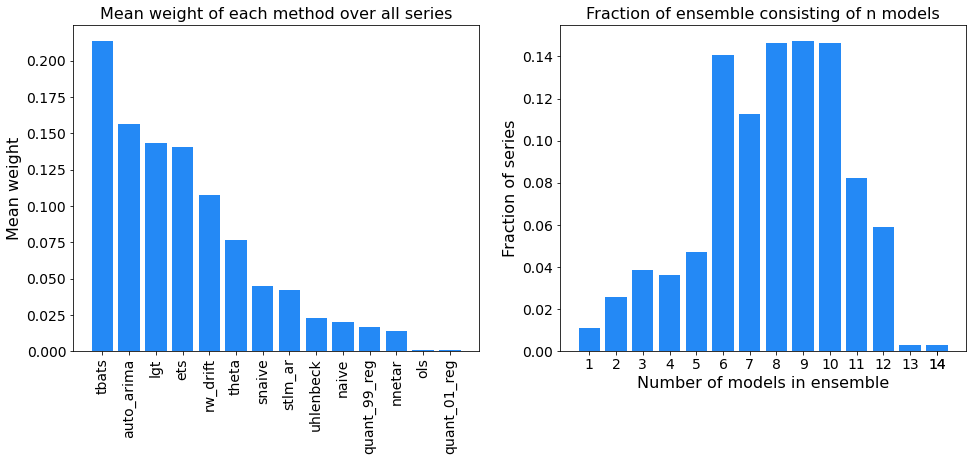}
    \caption{Left, a bar chart summarizing the mean weight each separate forecasting model was assigned by DONUT Fanciful over the entire test set. Right, the percentage of series for which DONUT Fanciful uses ensembles of a given size.}
    \label{fig:fanciful-combinations}
\end{figure}

In \autoref{fig:good-fanciful}, \autoref{fig:bad-fanciful}, and \autoref{fig:fanciful-combinations}, we illustrate how ``DONUT Fanciful'' weights the 14 available models. \autoref{fig:good-fanciful} shows an example of a good forecast with a MASE of 0.5, and the associated weights produced by a combination of ARIMA, TBATS, and Random Walk. \autoref{fig:bad-fanciful} shows an example of a bad forecast with a MASE loss of 4.4, with associated weights assigned to each model in the ensemble for the forecast. Here, the model used 0.8 of Seasonal Naïve and 0.13 of Exponential Smoothing, which did not yield a good result. It also looks like the series had a structural break that made it hard to forecast for the underlying methods.

In \autoref{fig:fanciful-combinations}, we find that the most popular method was TBATS with the highest mean weight out of any model, with ARIMA, LGT, and Exponential Smoothing relatively equal to each other at a slightly lower mean weight. OLS and Quantile regression with $\tau=0.01$ was virtually not used. We reason this might be because the Random Walk with Drift model of FFORMA is highly correlated with with both. Ornstein-Uhlenbeck is more popular than Naïve. It is also highly evident that ``DONUT Fanciful'' uses combinations of many different forecasting techniques at a high rate, as seen on the right in \autoref{fig:fanciful-combinations}; remarkably, ensembles consisting of 6-10 models seem most common.

\subsection{Performance Comparison to FFORMA model}
\label{sec:results-fforma-comparison}

Since the DONUT approach is primarily based upon the FFORMA model by \cite{montero2020fforma} with a goal of improving the approach by making fewer assumptions, it is uplifting that the performance has improved significantly. We make a simple comparison in \autoref{tab:fforma-improvment}. The ``DONUT Fanciful'' model improved the main metric of OWA by 0.008, which yields a statistically significant difference with an associated $p$-value of 0.6\%. This provides some evidence in favor of reducing the assumptions a model rests on.

\begin{table}
    \centering
    \footnotesize
    \singlespace
    \begin{tabular}{@{}lll@{}}
    \toprule
    $\mathbf{\Delta}$\textbf{OWA} & $\mathbf{t}$\textbf{-statistic} & $\mathbf{p}$\textbf{-value} \\ \midrule
    -0.008& -2.510   & 0.006$^c$  \\ \bottomrule
    \end{tabular}
    \caption{OWA improvement from FFORMA, with t-statistic and one sided p-value with $H_0$= FFORMA mean}
    \label{tab:fforma-improvment}
\end{table}

\begin{figure}
    \centering
    \includegraphics[width=0.8\textwidth]{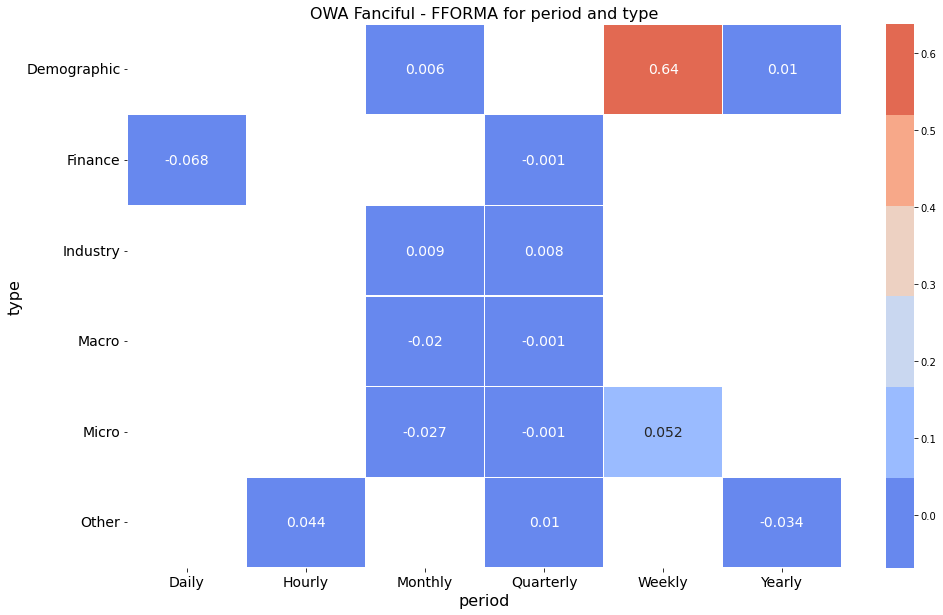}
    \caption{\textbf{The colored boxes have a} $\mathbf{p\textrm{-}value<0.05, \mu_0=mean(row_{box})}$: Our fanciful model has a mean OWA of 0.068 less than FFORMA for daily financial series, n=1,559. FFORMA is relatively stronger in the Demographic Weekly category, which contains only $n=24$ series.}
    \label{fig:type-period-vs}
\end{figure}

It is worth mentioning that a 0.008 OWA improvement would lead the sixth place competitor to take second place in the M4 Competition. As such, the improvement is noteworthy in the context in which it occurs. Even more so, one can look at what categories of series our net differs from FFORMA. In \autoref{fig:type-period-vs}, it is evident that our model performs better on the Financial Daily series by a margin of 0.07. FFORMA performs better in combinations such as Micro Weekly and Demographic Weekly. These two sets of categories, however, contain only 112 and 24 series, respectively.

\subsection{Feature-Importance Results}
\label{sec:results-feature-importance}

To discern whether the automatically extracted features of the autoencoder had any predictive power, we calculated the permutation feature importance scores for all features and compared them. We present an overview of the importance of each feature in \autoref{fig:feature-importance} and \autoref{tab:feature-importance} in the appendix. See \autoref{sec:methods-feature-importance} and \cite{breiman2001random} for details on how importances are calculated. In short, large increases in loss (decreased accuracy) means that the corresponding feature was more important for the prediction.

\begin{figure}
    \centering
    \includegraphics[width=0.9\textwidth]{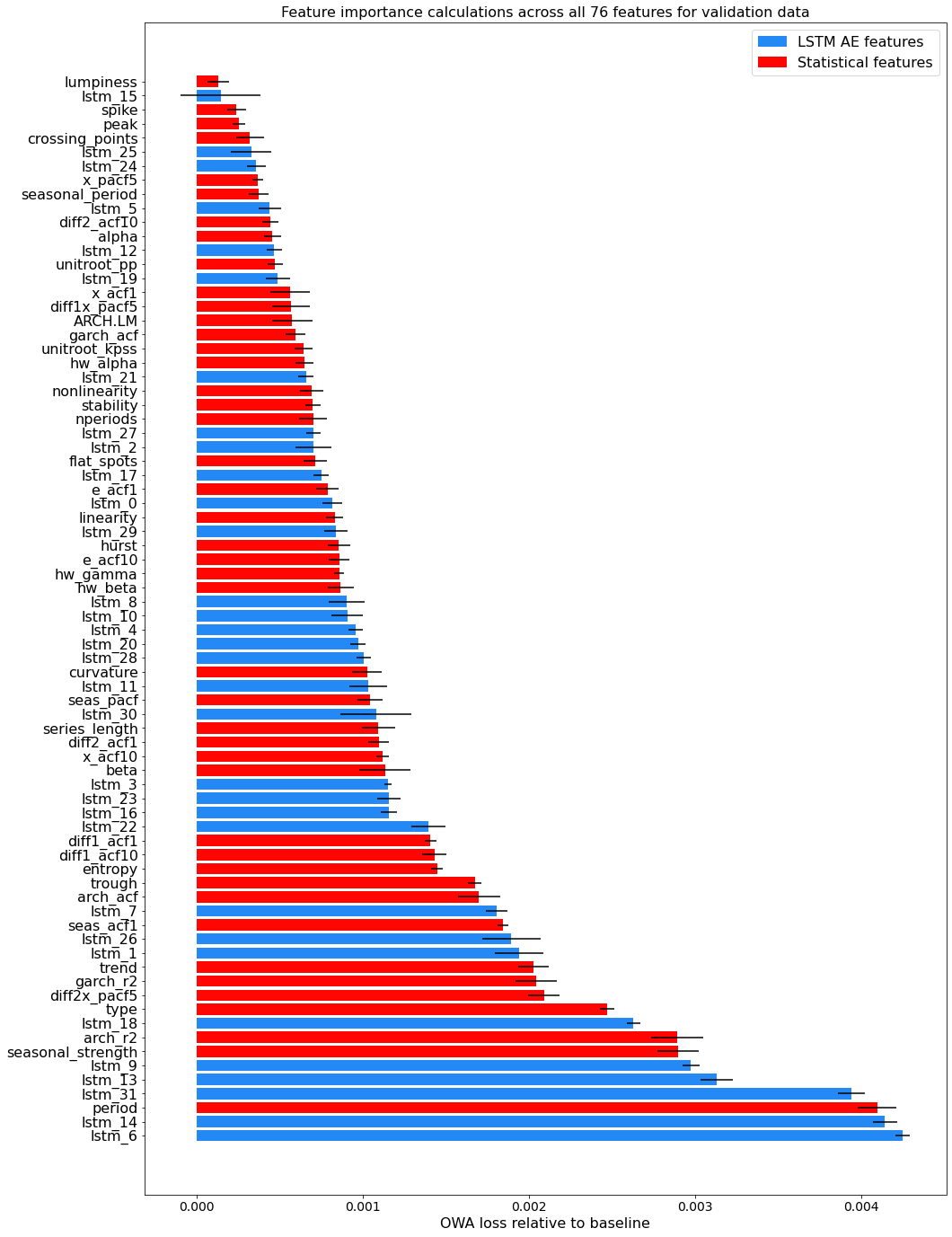}
    \caption{A visual representation of the feature importances for the 76 features our model uses. The features are listed in decreasing importance along the positive y-axis and x-values, which gives the deviation in OWA from the baseline when that feature is randomly permuted. The features generated by our Autoencoder are blue and the statistical features by \cite{montero2020fforma} are red. Several LSTM Autoencoder features have a high importance.}
    \label{fig:feature-importance}
\end{figure}

\begin{figure}
    \centering
    \includegraphics[width=0.9\textwidth]{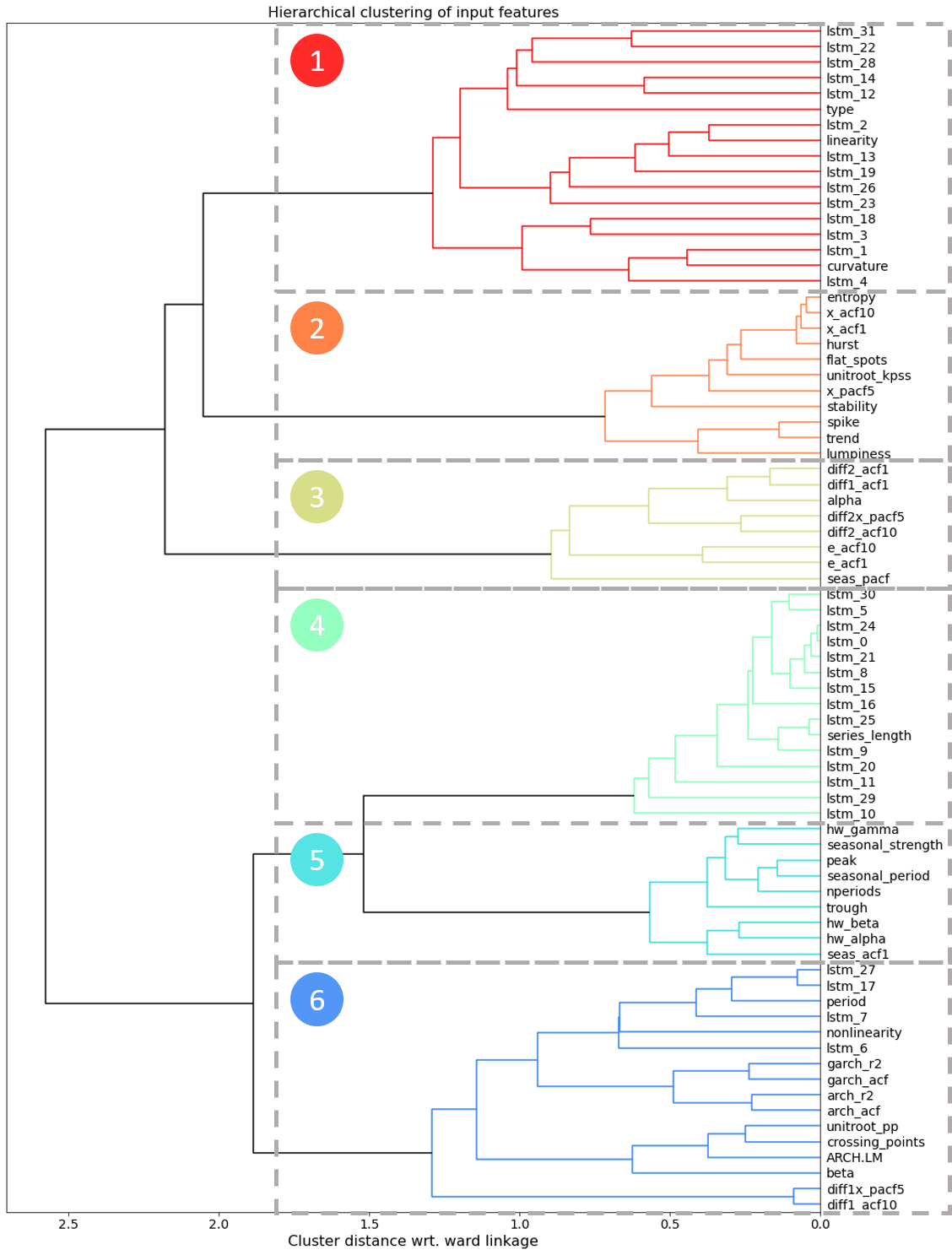}
    \caption{Hierarchical clustering of the correlation between all 76 features according to the ward's linkage. There is six distinct clusters of features formed. There is also a tendency towards clustering among the statistical features, and among the LSTM features.}
    \label{fig:feature-clustering}
\end{figure}

\begin{figure}
    \centering
    \includegraphics[width=0.9\textwidth]{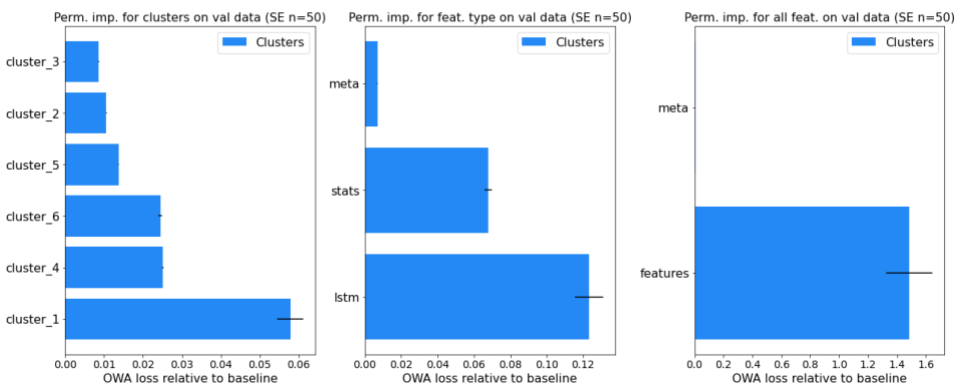}
    \caption{A plot showing the permutation importances (perm. imp.) when scrambling (1) all features in the different clusters, as defined in \autoref{fig:feature-clustering}, (2) including meta variables alone (type, period), all statistical features alone and all LSTM features alone, and (3) all features that are not the meta variables (type, period), from left to right, respectively.}
    \label{fig:feature-importance-3plots}
\end{figure}

Based on the clustering results in \autoref{fig:feature-clustering}, we also measured the permutation importance of all features within clusters to see what feature clusters are most important, as seen in the leftmost plot in \autoref{fig:feature-importance-3plots}. This is also compared to permutation importances for all statistical features used by \cite{montero2020fforma} together and all LSTM features proposed here together (middle plot in \autoref{fig:feature-importance-3plots}), as well as importance for all features used together (the rightmost plot in \autoref{fig:feature-importance-3plots}). We observe that the first cluster is the most important, with an OWA increase of $\sim 0.06$ when scrambled. This alone deteriorates the result significantly, as the resulting OWA draws the model several places lower on the leader-board. Interestingly, the top two clusters, Cluster 1 and Cluster 4, mainly consists of LSTM features. Also, the least important clusters, Clusters 2, 3, and 5, contain no LSTM features.

In the appendix, we have randomized all features five times, and provide $p$-values for all feature importances, sorted according to decreasing importance. Below, when we use the term ``importance,'' it refers to an increase in loss when scrambling the entire column of that feature.

It is evident that ``period'' is the most important statistical feature for prediction for the ``DONUT Fanciful'' weighting model. Seasonal effects are also important for our weight net, with ``seasonal\_strength'' being the second most important statistical feature, a result which coincides with the findings of \cite{makridakis1993m2}.

The LSTM features have a statistically significant high importance. Four out of the first five most important features are generated by the LSTM. Therefore, our weight net is able to extract information from these novel time series features, which is important for the weight it produces. ``lstm\_6'' and ``lstm\_14'' are the first and second most important features, even higher than that of ``period''. This is surprising in the sense that some LSTM values seem more important but we postulated that ``type'' and ``period'' was important for the network to make apt predictions.

It is also interesting to note that even the most important features increase the loss by a mere $~0.004$ in OWA loss. This means that removing one of the top features alone would not make the resulting OWA higher than what \cite{montero2020fforma} achieved. This can be viewed as a strength, and makes sense in the context of how the network is trained. We have employed a high degree of dropout regularization into the training, meaning that the network is trained under circumstances of high noise---resulting in a model highly robust to perturbations.

%% file: Sections/5_Discussion.tex
\section{Discussion}
\label{sec:discussion}

In this section, we reflect on the results, and present areas for further work. 

%\subsection{Something in the lines of "What our net thinks about methods and types of series"}

%E.g. Our net thinks that naive is the best forecasting method for XX\% of the financial times series.

\subsection{Reflections on the Results}

This paper sought to investigate two hypotheses: First (1), we hypothesized that reducing the number of assumptions on input features and relationships, models and outputs, as this could possibly lead to well-performing forecasting algorithms and increased accuracy. Second, (2) the hypothesis that state-of-the-art machine learning methods like LSTM Autoencoders can compress, extract and represent information, which is hard for humans to capture in as many as 42 human-defined statistical features.

\textbf{Hypothesis 1 finding: Fewer assumptions can lead to better results.}

We find our approach to out-compete other methods. The reduced assumptions in model building will likely lead to an increase in accuracy, as evidenced by the statistically significant OWA loss improvements and the DONUT model’s reliance on the novel Autoencoder features.

Our ensemble weighting model, built on the combination ensembling method, shows convincing results. With an OWA of 0.830 (see \autoref{tab:results-test-set}), our model beats 48 participants in the competition, only outperformed by \cite{smyl2020hybrid}, which uses an entirely different approach. We have also run a $t$-test on the result, with $H_0$ being that our true mean is equal to that of \cite{montero2020fforma}, which we reject with $p = 0.006$.

We find some evidence that adding more models to the ensemble improves the results regarding the model output. For example, \autoref{fig:fanciful-combinations} shows that the DONUT model's ensemble has a mean weighting of Ornstein-Uhlenbeck and a Quantile Regression with $\tau=0.99$, which is higher than some of the models already included in the FFORMA ensemble. This finding indicates that augmenting with diverse models can be beneficial. Moreover, in 27\% of cases, the weighting model uses 10 or more models. Thus, in a quarter of the series of the M4 Competition, the weighting model recommends a weighting that the FFORMA model could not possibly predict, given that its ensemble only contains nine models.

\textbf{Hypothesis 2 finding: Machine learning methods can find features humans cannot.}

There seems to be evidence to support the hypothesis that machine learning methods can find efficient representations of time series features that are hard for humans to do.

The features extracted by the Autoencoder represent statistically different information from those of the statistical features in several cases. For example, \autoref{fig:feature-importance} shows that, together with ``period,'' the LSTM features are among the most detrimental to scramble for the model; in this interpretation of feature ``importance,'' 4 out of 5 of the most ``important'' features, are generated by the LSTM. Even though each scrambling amounts to a 0.004 increase in OWA, the comparison of interest is the relative importance between the LSTM features and the statistical ones. Indeed, our results indicate that the LSTM features contain information as valuable, if not more valuable, than the statistical features. The fact that several models trained during the development of the DONUT approach demonstrated a reliance on many of the same LSTM features further supports the finding.

The \autoref{fig:feature-clustering} graph depicts a hierarchical clustering of feature correlation in the form of a dendrogram. For example, in the figure, one can see how the dendrogram clustered some LSTM features with each other and more distantly with the statistical features (see, e.g., Clusters 1 and 4 in \autoref{fig:feature-clustering}). This clustering might indicate that the LSTM features contain information not present in the statistical features alone. Moreover, some of the single most `important' statistical features, like ``lstm\_6'', ``lstm\_9'', ``lstm\_14'', are all present in different clusters, showing a high correlation distance to each other, indicating that they have also found different features that are all the while important.

\subsection{Further Work}

There is much testing we would have done with more time. This section offers some thoughts on possible improvements for the weighting model and the Autoencoder. The relationship between the post-facto optimal solution and the final prediction result could also offer appealing topics for further study. 

Firstly, one can do considerable optimization of the weighting model, DONUT. We wanted to ensemble many of our fully trained models to create an ensemble of ensembles. We believe the meta-ensemble would be compelling, since we observed different ways of failing and utilizing the features between the weighting Neural Nets trained with slightly different hyper-parameters. Alas, time would not allow this.

We would also have liked to use feature selection to remove features that did not contribute sufficiently to the weighting model's predictive power. From \autoref{fig:feature-importance}, it is evident that ``lumpiness'', ``lstm\_15'' and ``spike'' is direct noise to the weighting entity. As such, removing these could help in the weight nets predictive power.

We also have enticing possibilities for further work on the Autoencoder. For example, we would have liked to reduce the number of features to make each dimension contain more predictive power in the embedding vector. This dimensionality reduction seems an even more attractive option after the results found in \autoref{fig:feature-importance}. 

We think there are more effective relationships to study between the ex post facto optimal solution and the actual predictive performance. Indeed, the ex-post facto solution offers an upper bound for the weighting entity, as seen in \autoref{eq:T-loss}. Compared with the forecast selection's ex-post-facto approach, the movement of this bound could lead to some fascinating results.

We would have liked to train on more data, and extend the model with relevant panel data. One could also add augmented data specifically aimed at increasing performance, in which the ensemble cannot predict well, or for the specific data categories which are underrepresented. We would also have liked to try our ensemble forecasting approach in real-world applications, such as predicting epidemic data or prices.

%% file: Sections/6_Conclusion.tex
\section{Conclusion}
This paper aimed to reduce assumptions about the input and output of a combination ensemble with a DONUT procedure (DO Not UTilize human assumptions). We also wanted to investigate whether LSTM-generated time-series features contain more crucial information than traditional statistical features. Our reduction of input assumptions through auto-generated features as part of an ensemble model outperforms the statistical feature-based ensemble FFORMA by \cite{montero2020fforma}. Furthermore, we find that our weighting model uses both LSTM features and statistical features to achieve this. Minor indications point in the direction of LSTM superiority over statistical features alone. In a correlation dendrogram analysis, it is evident that LSTM features are indeed different from one another, and form two clusters dissimilar from most standard statistical features. We also find that increasing output space by adding available models is something the weighting model learns to use, there partly explaining the accuracy gains.

We also quantified the superiority of model combination compared to model selection a posteriori through linear optimization on the M4 dataset.

Lastly, our findings indicate that classical statistical time series features, such as trend and seasonality, alone does not capture all relevant information for forecasting a time series. On the contrary, our novel LSTM features seem to contain significantly more predictive power than the statistical ones.

%% file: Appendices/Theory.tex
\section{Theory}

\subsection{Findings From Previous M Competitions}
\label{app:m-findings}
\begin{table}[ht]
\centering
\tiny
\begin{tabular}{l}
\hline 
M1, 1982 \\ \hline

\tabitem The accuracy measure utilized can change the ranking of the different methods            \\
\tabitem Performance of the various methods depend upon the length of the forecasting horizon \\
\tabitem Accuracy of combinations of methods outperforms on average the single \\ \qquad methods alone and does well in comparison with other methods \\
\tabitem Given an ensemble of methods, the simple average over them outperformed \\ \qquad the more complex average based on covariates, though that still did well \\ \hline
M2, 1993 \\ \hline
\tabitem Good and robust performance of exponential smoothing methods. \\
\tabitem Less randomness leads to better relative accuracy of the more sophisticated methods \\
\tabitem The greatest improvement in forecasting accuracy came from measurement \\ \qquad and extrapolating the seasonality of the series$^\dagger$ \\ \hline
M3, 2000 \\ \hline
\tabitem Statistical and sophisticated methods do not necessarily \\  \qquad produce more accurate forecast than simple ones$^\dagger$ \\
\tabitem Accuracy of combinations of methods outperforms on average \\ \qquad the single methods alone and does well in comparison with other methods$^\dagger$      \\ \hline
\end{tabular}

\caption[M-Competitions Summary]{Summarized M competition findings. $^\dagger$The most relevant findings for this paper are marked with a dagger.}
\label{tab:M_competitions}
\end{table}

\subsection{Statistical Forecasting Models}
\label{app:new-models}

In this section, we explain the workings of five models we have added to the existing nine models used by \cite{montero2020fforma}, and some rationale for adding each model to the set of available models $\mathcal{M}$.

\subsubsection{Ordinary Least Squares}

Ordinary Least Squares (OLS) is a method that permeates the statistical landscape. First introduced by \cite{gauss1823theoria}, OLS is a method of minimizing the squared residuals between a dependent variable and a linear combination of independent variables. The Gauss-Markov theorem as presented in \cite{greene2018econometric} states the following.

\textit{``In the linear regression model with given regressor matrix $\boldsymbol{M}$, (1) the least squares estimator $\boldsymbol{b}$ is the minimum variance linear unbiased estimator of $\boldsymbol{\beta}$ and (2) for any vector of constants $\boldsymbol{w}$, the minimum variance linear unbiased estimator of $\boldsymbol{w}'\boldsymbol{\beta}$ is $\boldsymbol{w}'\boldsymbol{b}$.''}

\cite{greene2018econometric} presents five requirements for the OLS estimator to have the qualities of a Best, Linear, Unbiased Estimator (BLUE), a commonly desired property of an estimator. These are as follows. A1. The regression model is linear in its parameters. A2. There is a random sampling of observations. A3. A conditional mean of zero. A4. There is no multi-collinearity (or perfect collinearity). A5. Spherical errors: There is homoskedasticity and no auto-correlation.

Thus, we think that given that a time series follows the above properties and that there is a linear relationship between time (our independent variable in the case of the M4 univariate series) and the dependent variable, OLS could be an excellent model to add to our ensemble. Furthermore, since several of the statistical features we use tests for homoskedasticity, auto-correlation, and orders of integration, we stipulate that the neural net is provided the necessary information to weight OLS appropriately when these criteria take on values that allows OLS to be BLUE. For instance, the Kwiatkowski–Phillips–Schmidt–Shin test (KPSS), which tests for trend-stationary time series \cite{kwiatkowski1992testing}, is one of the available features. Thus, the weighting model could utilize the linear model when the KPSS feature takes appropriate values. In principle, the model could then deploy OLS with reasonable confidence that the conditional mean of the error term is 0.

The OLS univariate linear regression-based model would perform well for a process $y_t$ defined as

\begin{align}
    y_t = \alpha + \beta t+ \epsilon_t,
\end{align}

where $y_t$ is the value of the process, $\alpha$ a level term, $\beta$ the linear relationship coefficient between the time $t$ and the dependent variable, and $\epsilon_t$ is a normally distributed error term with mean 0. The literature defines a model estimating the process in \autoref{eq:ols} as

\begin{align}
    \label{eq:ols}
    \Hat{y}_t &= \Hat{\alpha} + \Hat{\beta} t, \\
    \text{where}\quad \Hat{\beta} &= \frac{n\sum t y_t-\sum t \sum y_t}{n \sum t^2-(\sum t)^2}, \\ \text{and} \quad \Hat{\alpha} &= \Bar{y} - \Hat{\beta} \Bar{x}.
\end{align}

In the above, the hats above the parameters, e.g., $\Hat{\alpha}$, denotes estimated parameters. In a multivariate case, one could add more regressors and more advanced methods with dummy variables to model, e.g., seasonality and structural breaks. Other models in our ensemble try to utilize these more advanced forecasting techniques.

\subsubsection{Ornstein-Uhlenbeck Process}

An Ornstein-Uhlenbeck (OU) process is a stochastic differential model representing mean-reverting processes \cite{schobel1999stochastic}. We define a simplified model for the dependent variable as

\begin{equation}
    \label{eq:ornstein}
    dS = \gamma(m-S)dt + \sigma dZ,
\end{equation}

where $S$ is the dependent variable, $m$ is the time-independent mean, $dt$ is the change in time and $dZ$ is a Wiener process \cite{bibbona2008ornstein,wiener1976collected}. One often interprets $\gamma$ as the reversion velocity, i.e., how fast the process returns to the steady-state once a shock or deviation occurs. One can analytically solve the above stochastic differential equation. However, it is more efficient to estimate it using numerical methods for computational approaches using the above model.

We chose to add the OU process as we thought it possible for the weighting model to identify mean-reverting time series, and the OU model will complement the existing models, cf. the discussion in \autoref{sec:theory-ensemble-forecasting}. E.g., researchers often model commodity prices as mean-reverting processes, and there are both financial theory and empirical evidence for the existence of mean-reverting assets. For example, \cite{bessembinder1995mean} finds that 44\% of a typical spot oil price shock, in expectation, reverses over the following eight months. Contextualizing this, given that our task for monthly series is to predict 18 time-steps into the future, adding a mean-reverting process could account for such a shock given that the weighting model can identify some probability of mean reversion in a series. Furthermore, the financial theory argument is, in short, based on supply and demand, and the tendency of supply and demand to increase and decrease respectively at a higher price and thus shock adjusting prices (and vice versa for low price shock).

Commonly, practitioners find OU parameters by using the maximum likelihood method. We mention that we have adjusted $\sigma$ in our final implementation since the Wiener Process explains the forecast $S$ in the short term to a more considerable extent. However, it is desirable for the model to predict signal rather than noise, and have thus opted to reduce the noise in the Ornstein-Uhlenbeck process, as we stipulate that it will increase the speed of learning how to act if and when mean reversion occurs.

\subsubsection{Local and Global Trend Model}

The Local and Global Trend model (LGT) is perhaps the least known methods of the ones chosen to augment the ensemble. LGT, as presented by \cite{ng2020orbit}, is a statistical forecasting model combining different time series attributes to generate forecasts,

\begin{align}
   & y_t = \mu_t + s_t + \epsilon_t \\ \label{eq:measurement}
    &\mu_t = l_{t - 1} + \xi_1b_{t - 1} + \xi_2l^{\lambda}_{t-1}\\ \label{eq:transition}
    &\epsilon \sim \textrm{Student}(0,v_0,\sigma)\\
    &\sigma \sim \textrm{Gauchy}(0, \gamma_0),
\end{align}

where $l_{t}, \xi_1b_{t}, \xi_2l^{\lambda}_{t-1}, s_t, \epsilon_t$ are modeling level, local trend, global trend, seasonality, and error term, respectively. $\gamma_0$ is a data driven scalar. As such, the model follows a triple exponential smoothing form in the updating process, which has proven effective in practical forecasting applications \cite{siregar2017comparison}.

The model is a state-space model and is thus part of the large body of work in this field \cite{ng2020orbit}. State-space models have in common that they try to explain a dependent variable by another state variable, of which it has its name. In literature, these are often called the measurement equation and the transition equation, where the first is the observed dependent variable while the other tracks the change or transition of the state \cite{brooks_2019}. In the specific example of LGT, \autoref{eq:measurement} and \autoref{eq:transition} would be the measurement and transition equation, respectively.

The main reason for the addition of LGT is the promising results shown by \cite{ng2020orbit}. Out of the models compared in the paper, LGT is the number one method for predicting M3 Competition data, significantly beating benchmarks such as sARIMA. On M4 weekly data, it underperforms compared to some methods but outperforms the remaining two other data sets analyzed in the paper. Another reason for the addition of LGT is that a state-space model seems plausible that some time series are well described. It would thus be interesting to analyze what weight, and in what situations such a model would be given weight, by the weighting model.

Lastly, a Python implementation of LGT is freely available in the GitHub package \href{https://orbit-ml.readthedocs.io/en/stable/tutorials/lgt.html}{Orbit}, and as such, the model is efficiently implemented and added to the ensemble, despite being reasonably complex.

\subsubsection{Quantile Regression}

Quantile regression is a method that resembles the classic OLS regression, except for estimating the conditional median of the dependent variable instead of the mean \cite{brooks_2019}. 

Quantile regression, first developed by \cite{koenker1978regression}, offers a way of handling complex relationships between a dependent variable and regressors using conditional quantiles. For example, the median is less sensitive to outliers than the mean, and quantile regressions offer a more versatile distribution measure over possible values than standard OLS. Another strength is that it is non-parametric and does not make any assumptions about the variable distribution. Furthermore, we can adjust the model to forecast the quantiles $\tau$ in which we are interested.

\cite{brooks_2019} defines quantile regression, for a given quantile $\tau$, as a representation of the model to estimate $\hat{\beta}$ as

\begin{align}\label{eq:quant_reg}
    &\hat{\beta} = \textrm{argmin}_{\beta} \Big( {\beta}(\sum^{N}_{i:y_i > \beta x_i}\tau|y_i-\beta x_i| + \sum^{N}_{i:y_i < \beta x_i}(1-\tau)|y_i-\beta x_i| \Big) \mid \\
    &  Q(\tau) = \textrm{infimum }y: CD(y) \geq \tau,
\end{align}

where $CD(\cdot)$ is the cumulative distribution, and infimum is the greatest lower bound such that the inequality is satisfied. In our case, the regressor $x_i$ is as in \autoref{eq:ols} i.e. the time step $t$.

In our ensemble, $\mathcal{M}$, we have included quantile regressions the two values of $\tau$, $\tau_{.99}=.99$ and $\tau_{.01}=0.01$. We chose these levels for the quantile regressions to provide the weighting model with a larger space of possible forecasts to choose from, again cf. the discussion of model diversity in \autoref{sec:theory-ensemble-forecasting}. The $\tau_{.99}$ level regression provides the weighting model with the possibility of predicting steeper growth than any of the other models alone. The $\tau_{.01}=0.01$ level regression, on the other hand, makes it possible to add hard dampening of the forecasted growth of a series. Both cases rely on the model learning a relationship between the input features and series where such increased or decreased growth is beneficial.

\subsection{Artificial Neural Networks}
\label{app:ann}

\subsubsection{The Basics of Artificial Neural Networks}

An Artificial Neural Network (NN) is a subset of machine learning methods that are given their name by mimicking the human brain. In our full ensemble model, we use two NNs. One is the Fully Connected NN that produces the weights for the different forecasting models $m\in\mathcal{M}$, thus producing a prediction $\mathbf{\Hat {y}_{t+h}}$. The other is the Long Short-Term NN which extracts time-invariant features from the time series, which we discuss in more detail in \autoref{sec:lstm}).

A NN performs non-linear transformations from input to output and can, in theory, approximate any real-valued function from one finite-dimensional space to another to any desired degree of accuracy. Provided there is at least one hidden layer, a non-linear squashing function is present, and sufficiently many hidden units are available \cite{hornik1989multilayer}. A NN works by iteratively using matrix multiplication (linear transformation), importantly followed by a non-linear function \cite{Goodfellow-et-al-2016}. \autoref{eq:non-lin} shows a single non-linear transformation layer $\mathbf{L}$, where $g$ is a non-linear activation function, $\mathbf{A}$ is a linear transformation, and $\mathbf{b}$ is the bias. There is a difference in the complexity a linear, and non-linear model can learn, for instance, the XOR function, which is infeasible for a linear function but trivial for several non-linear models \cite{minsky2017perceptrons}.

\begin{equation}
\label{eq:non-lin}
    \mathbf{L} = g(\mathbf{A}^T \mathbf{x}+\mathbf{b})
\end{equation}

One can depict a NN as a graph, and, more formally, it is the iterative application of the above function for an arbitrary number of steps, with an arbitrary number of parameters at each step (this flexibility makes the model susceptible to being over-parameterized, which we briefly discuss in \autoref{sec:theory-early-stopping}). For instance a neural network with one hidden layer could be written as $y_{dep} = L_{out} \circ L_h \circ L_{in}(x_{ind})$. The first layer, the input layer, is provided data, which is passed through the network, layer by layer. At the last layer, $L_{out}$ creates the output with a fitting activation that is application-specific ( e.g., classification or regression). The output then flows to a loss measure (see \autoref{sec:loss_functions}), which updates the parameters in the network (often called weights), backward through the layers in a process called backpropagation \cite{Goodfellow-et-al-2016}. Note that if one were to remove $g$ from \autoref{eq:non-lin}, the complexity increase from iteratively adding layers would be removed as $\mathbf{A_1}\mathbf{A_2}...\mathbf{A_n}\mathbf{x} = \mathbf{A}\mathbf{x}$, results in a linear transformation, where $\mathbf{A_i}$ are all linear transformations. 

A Fully Connected Neural Network, as our weighting model is an example of, is an NN where all layers are fully connected, i.e., there is an edge between each node in one layer to all nodes in the next ($nm$ connections between a layer with $n$ and $m$ nodes).

\subsubsection{Gradient Descent}

Gradient descent is a way of fitting NNs by reducing the value of $f(x)$ by moving a small negative direction of the derivative. \autoref{eq:gradient_descent} shows how gradient descent is deduced as in \cite{Goodfellow-et-al-2016}. Since the relationship

\begin{align}
    \label{eq:gradient_descent} 
    &f(x+\epsilon)\approx f(x)+\epsilon \dfrac{df}{dx},
\end{align}

holds at sufficiently small $\epsilon$, the derivative helps minimize a loss function as it shows what direction would cause a decrease in the loss (i.e., an increase in performance or accuracy). This observation leads to the relationship

\begin{align}
    &\rightarrow f(x - \epsilon \cdot \text{sign}(\dfrac{df}{dx})) f(x), 
\end{align}

for $\epsilon$ small enough. Therefore, we can minimize the loss by moving in small increments in the opposite direction of the derivative.  \cite{cauchy1847methode} originally termed this process gradient descent.

Gradient descent is the primary step of backpropagation, where each weight is updated recursively backward through the model, layer by layer. Gradient descent can also be generalized for multi-input cases, where the aim would be to move in the direction of the steepest descent in the multidimensional space of the loss function. In the multi-variable case, we would get

\begin{align}
    \mathbf{\theta'} = \mathbf{\theta} - \epsilon \nabla_\theta \mathcal{L}(\mathbf{\theta}),
\end{align}

where $\epsilon$ is the learning rate and $\nabla_\theta \mathcal{L}(\mathbf{\theta})$ is the vector containing all partial derivatives of the loss function $\mathcal{L}$ with respect to the model parameters $\theta$ \cite{Goodfellow-et-al-2016}.

\subsubsection{Hyperparameter Tuning}
\label{sec:hyperparam-tuning}

Modern machine learning models have increased exponentially in complexity over the last couple of decades, and with it, an exponentially increasing space of hyperparameters to search over \cite{li2017hyperband}. Unfortunately, there are, in general, no methods that guarantee optimal hyperparameters except for a brute force search, for most hyperparameters \cite{snoek2012practical}. We present a discussion about how we can solve this vital issue in \autoref{sec:method-hyperparameter-search}.

\subsubsection{Early Stopping}
\label{sec:theory-early-stopping}

\begin{figure}[t]
    \centering
    \includegraphics[width=0.6\textwidth]{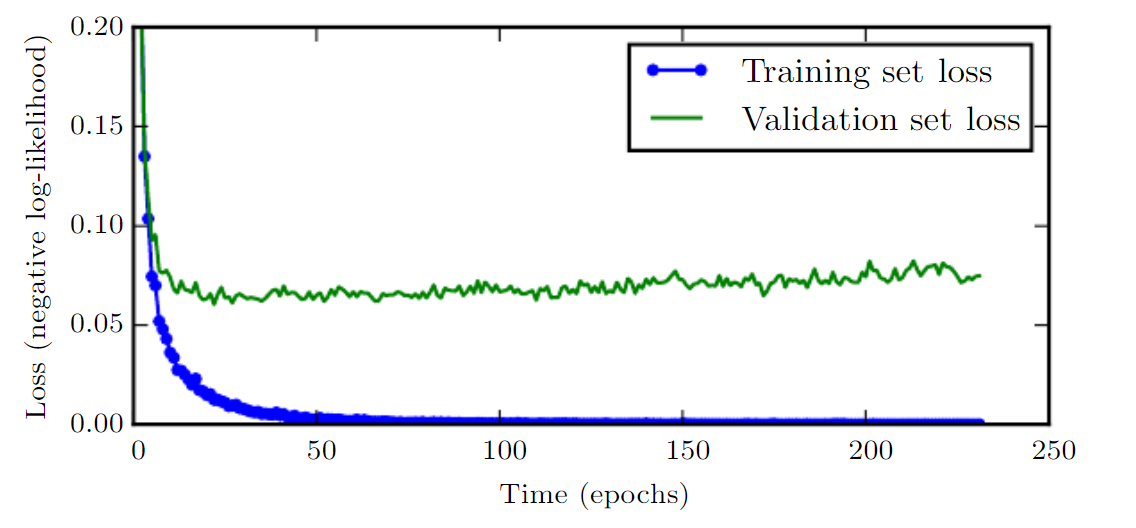}
    \caption[Overfitting Illustration]{Illustration of the tell-tale signature of a model that has been overfitted. The defining factor is that the training loss falls (blue line) while the validation loss (green) starts to increase. (Source: \cite{Goodfellow-et-al-2016})}
    \label{fig:goodfellow-early-stop}
\end{figure}

Large models with a sufficiently large number of parameters can suffer from overfitting, meaning that the model stops learning general relationships between input and output and instead learns to fit the noise of, or memorize, the training set. We can observe this phenomenon by a deviation in training and validation loss, exemplified in \autoref{fig:goodfellow-early-stop} \cite{Goodfellow-et-al-2016}. \cite{Goodfellow-et-al-2016} argue that the best regularization technique for NNs is to stop the training at the point where the validation loss is at its lowest, known as \textit{early stopping}. This regularization technique requires using some of the training data as a validation set, reducing the size of the training set. This limitation can be overcome by retraining on the entire training set using the exact specifications as found previously with the optimal hyperparameters and stopping point \cite{Goodfellow-et-al-2016}.

%% file: Appendices/Methods.tex
\section{Methods}

\subsection{Automated Hyperparameter Search}
\label{sec:method-hyperparameter-search}

One drawback with Deep Neural Nets is the large number of hyperparameters that must be set just right to achieve good results.

\begin{figure}[H]
    \centering
    \includegraphics[width=1.0\textwidth]{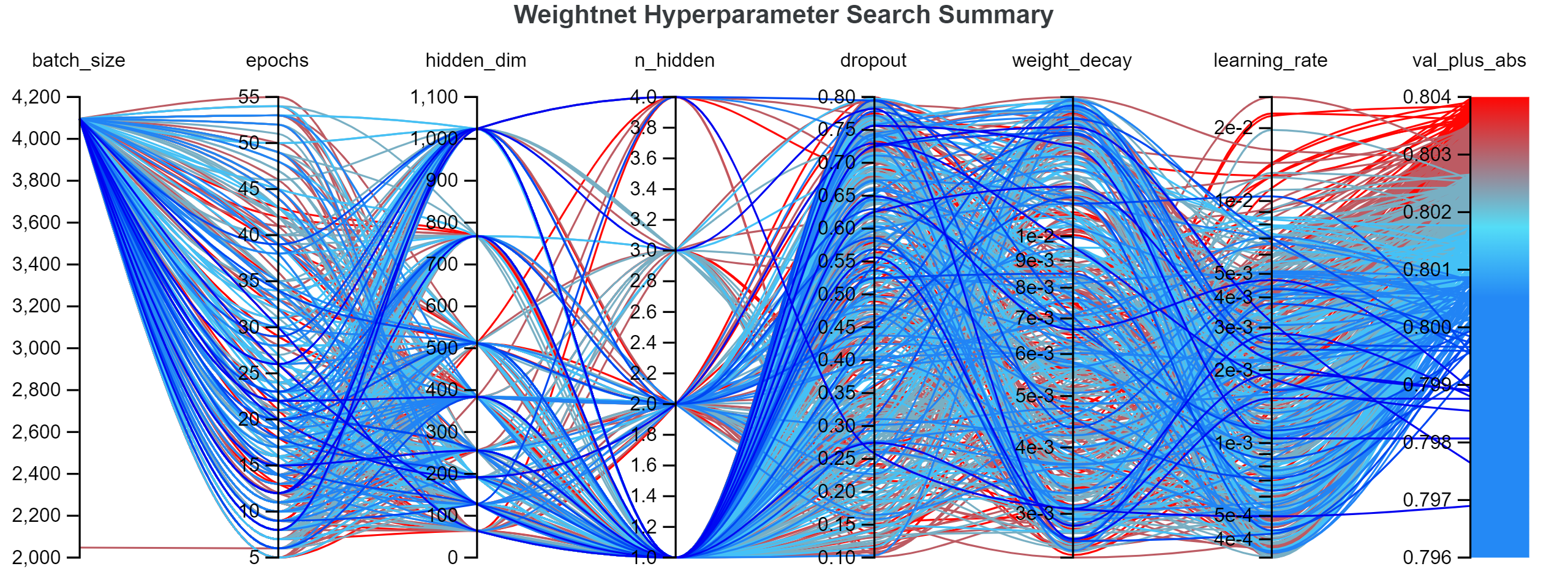}
    \caption[Hyperparameter search graph]{Summary over multiple runs with different hyperparameters and how they relate to final validation loss. Each line from left to right is one full training of the model with a unique set of hyperparameters characterized by where the line crosses the different lines for each hyperparameter. Deep blue lines achieved the best validation loss, and deep red is the worst.}
    \label{fig:sweep-flowchart}
\end{figure}

In \autoref{fig:sweep-flowchart}, the different combinations of hyperparameters and their corresponding validation loss can be seen. The most striking result from this plot is the observation that the model seems to prefer one hidden dimension, as well as considerable degrees of regularization in the form of dropout.
Other conclusions to draw from the analysis are that, in general, larger batch size and a lower learning rate boosts performance.

\subsection{LSTM Autoencoder Interpretation}
\label{app:ae-interpretation}

\subsubsection{Reconstruction Results and Interpretation}
\begin{figure}[H]
    \centering
    \includegraphics[width=0.9\textwidth]{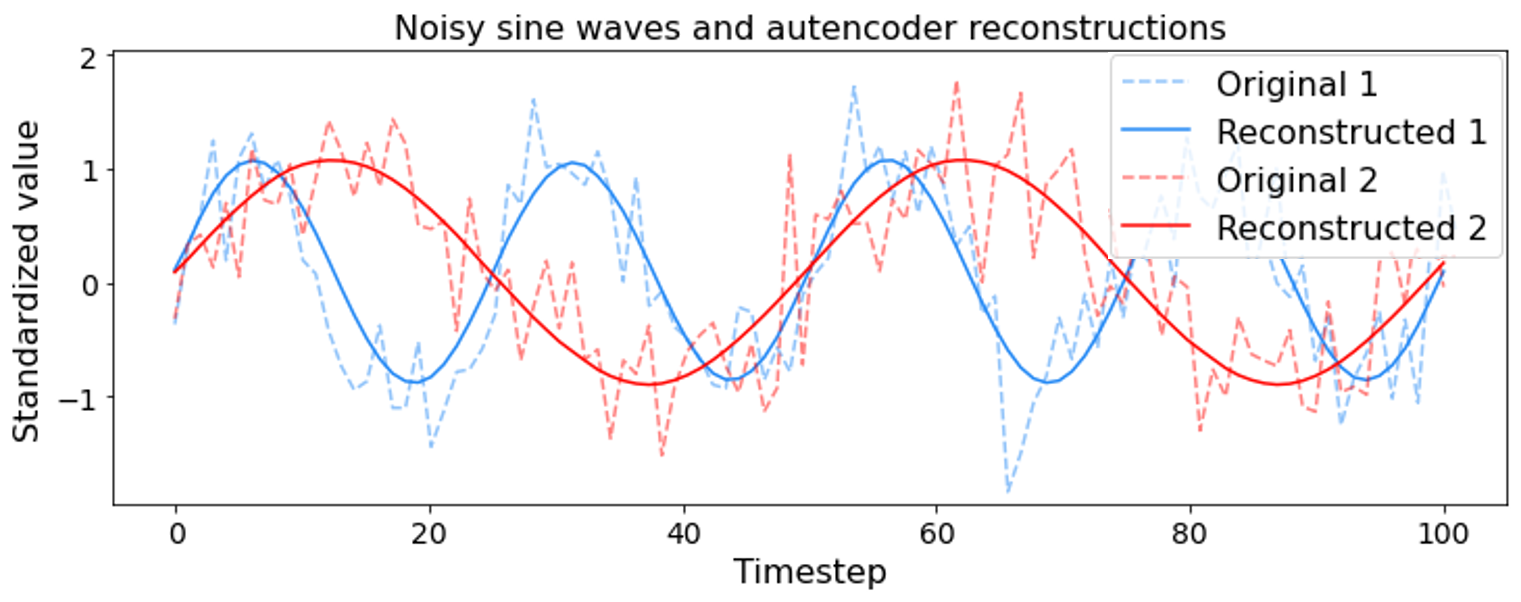}
    \caption{Two different noisy sine-waves and their reconstructions. The noise is clearly filtered out and only the smooth sine-waves remain. The embedding dim used to represent recreate the sine curve is of length 4, while there are 100 time steps, ergo the recreation must be based on a compressed abstraction of sine}
    \label{fig:sine-ae-recon}
\end{figure}

To validate that the network did indeed learn something, we constructed several different synthetic datasets, one of which can be seen in \autoref{fig:sine-ae-recon}. The dashed blue and red line is two different sine waves with random noise added. After training on the M4 data, the Autoencoder's reconstruction is the blue and red lines. It has never seen these lines before. Still, it filters out the noise to a very high degree and captures the essence of the series. Synthetic data was used for this because it allows one to decide what percentage of the data is signal and noise. Furthermore, in this case, the embedding dim is of length 4, while there are 100 time steps. As such, this Autoencoder embedding dimension must contain an abstraction of sine, which the Decoder can construct at least 25 times its input size.

\begin{figure}[H]
    \centering
    \includegraphics[width=\textwidth]{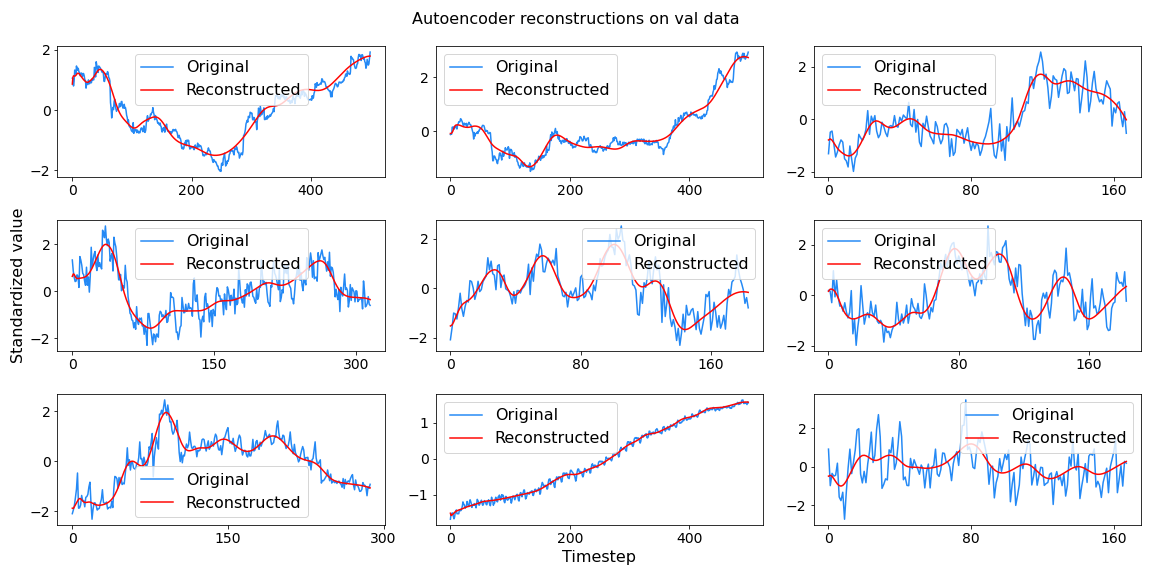}
    \caption{Original and reconstructed time-series from the M4 dataset of the validation partition, i.e. out-sample data. The autoencoder is able reconstruct many different types of time-series of different lengths, only based on a feature vector of length 32.}
    \label{fig:valdata-ae-recon}
\end{figure}

In \autoref{fig:valdata-ae-recon} some examples of our actual AN is depicted. The reconstructions are done on the validation part of the M4 dataset, meaning that these are graphs the AN has never seen. Here, the Autoencoder is trained to reconstruct the last 1 000 data-points of longer series, or the full length for shorter series (in the example, all happen to be of length 400 or shorter). Once again, the network is, to a high degree, able to capture the essence of the time series, despite never having seen these series before and is limited to representing the whole series in a vector of length 32 ($<< 1000$, i.e., the length of each time-series).

\subsection{Linear program}
 \label{linear-program}
 $E_{loss}(\mathcal{M,\mathbf{y_{t+h}}})=$

\begin{eqnarray*}
 \textrm{Sets} &  \mathcal{M}  & \textrm{Models in ensemble} \\
 &  T  & \textrm{Forecast horizon} \\
  &&\\
 \textrm{Parameters} & b_{ij}, (i,t) \in \mathcal{M} \times T & \textrm{Matrix parameter representing} \\
    && \textrm{model $i$ forecast at $t$}\\
 & y_{t}, t \in T & \textrm{Matrix parameter representing actuals} \\
 &&\\
 \textrm{Variables} 
 & x_{i}, i \in \mathcal{M} & \textrm{The weight of each model} \\
 & z^{+}_{t},z^{-}_{t}, t \in T & \textrm{Variable for modeling absolute value distance} \\
  &&\\
  \textrm{Minimize} 
 & \displaystyle\sum_{t \in T} z^{+}_{t} +z^{-}_{t} & \textrm{Minimize total distance from actual} \\
  &&\\
 \textrm{Subject to:}
   & y_t - \displaystyle\sum_{i \in \mathcal{M}} x_{i}b_{it} \leq z^{+}_{t}, \forall t \in T & \textrm{The actual value is larger than the forecast} \\
   &  - y_t + \displaystyle\sum_{i \in \mathcal{M}} x_{i}b_{it} \leq z^{-}_{t}, \forall t \in T & \textrm{The actual value is smaller than the forecast} \\
   &\displaystyle\sum_{i \in \mathcal{M}} x_i = 1 & \textrm{Weighted sum equals 1} \\
   &  0 \leq x_{i} , \forall (i) \in \mathcal{M} & \textrm{Each weight non-negative}
\end{eqnarray*}

\begin{figure}
    \centering
    \includegraphics[width=\textwidth]{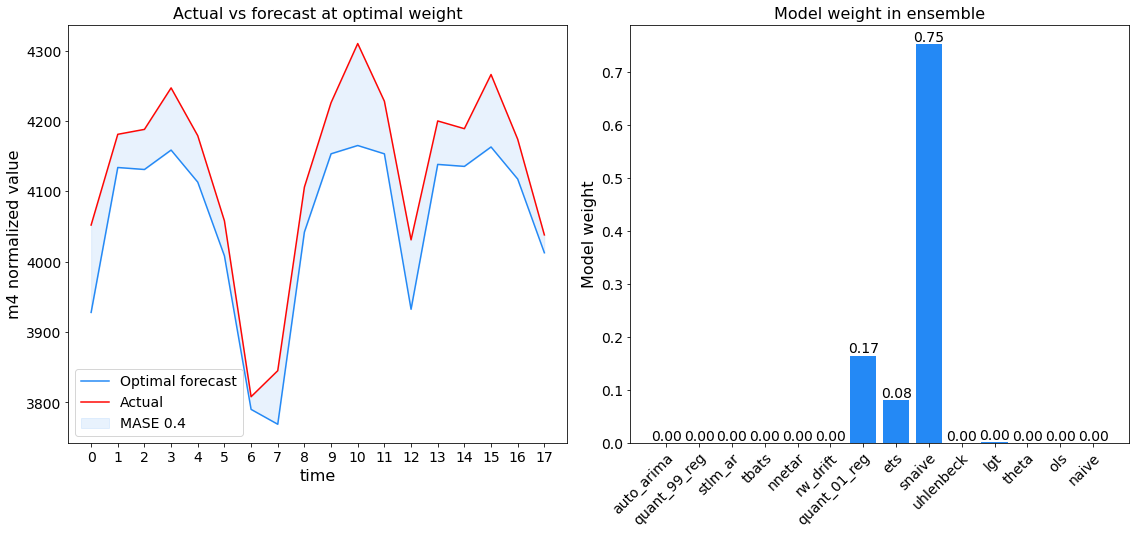}
    \caption{This is a plot where the optimal linear program solution clearly uses many forecasts and is successful. It is however not possible for the weight net to achieve a better result than 0.3 MASE given $\mathcal{M}_{14}$.}
    \label{fig:optimal-plot-good}
\end{figure}

\begin{figure}
    \centering
    \includegraphics[width=\textwidth]{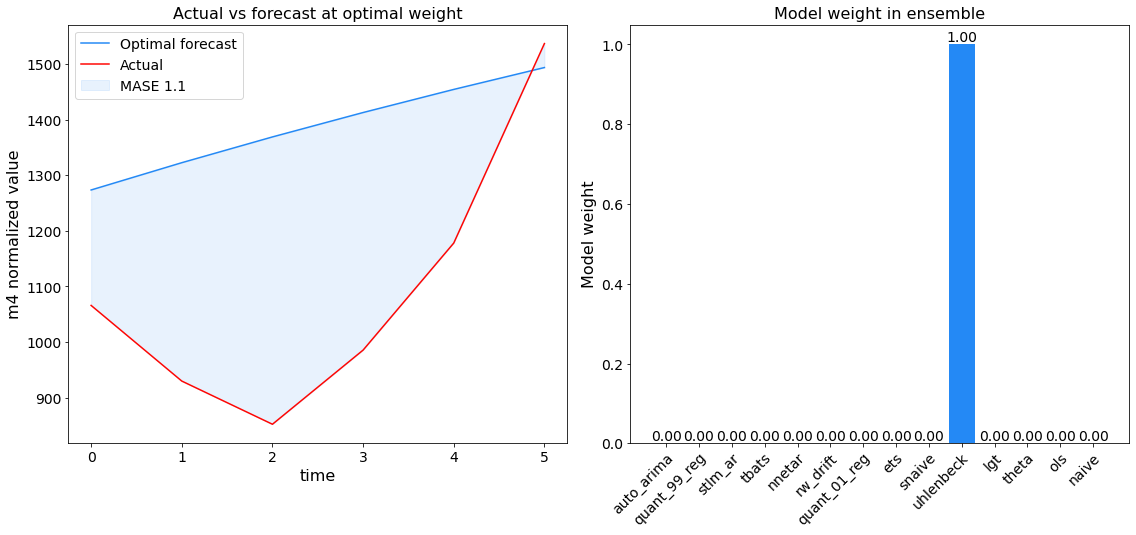}
    \caption{A optimal forecast where the linear program solution is no where near the actual times series. Furthermore, just a single forecast was used, implying that all other methods were even further from ground truth.}
    \label{fig:optimal-plot-bad}
\end{figure}

\begin{figure}
    \centering
    \includegraphics[width=\textwidth]{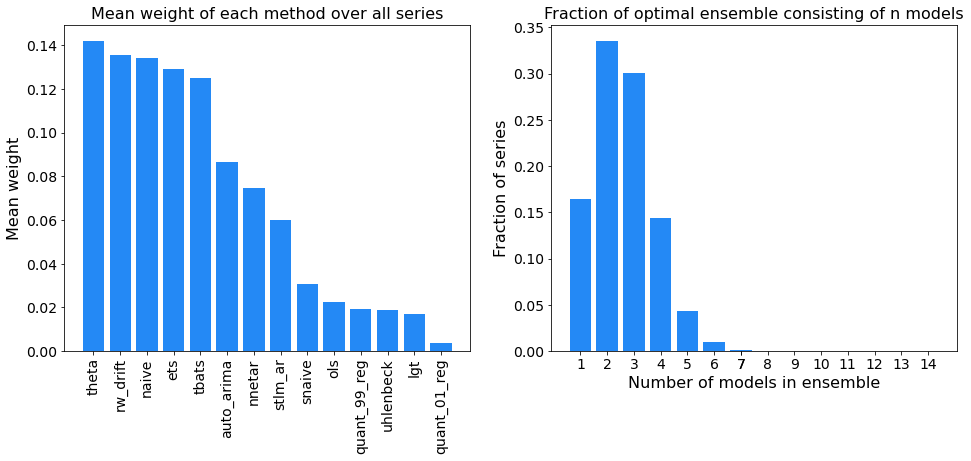}
    \caption{A display of the ex post facto best weight of the models in our 14 ensembles on our validation set. To the left is a display of mean weights for all models. Our newly implemented models are given a relatively low weight ex post facto. The right figure shows that 2-3 models ensemble is the most common combination. 4 models and more add up to around 20\% of all ensembles. }
    \label{fig:opimization-14-models}
\end{figure}

\begin{figure}
    \centering
    \includegraphics[width=0.9\textwidth]{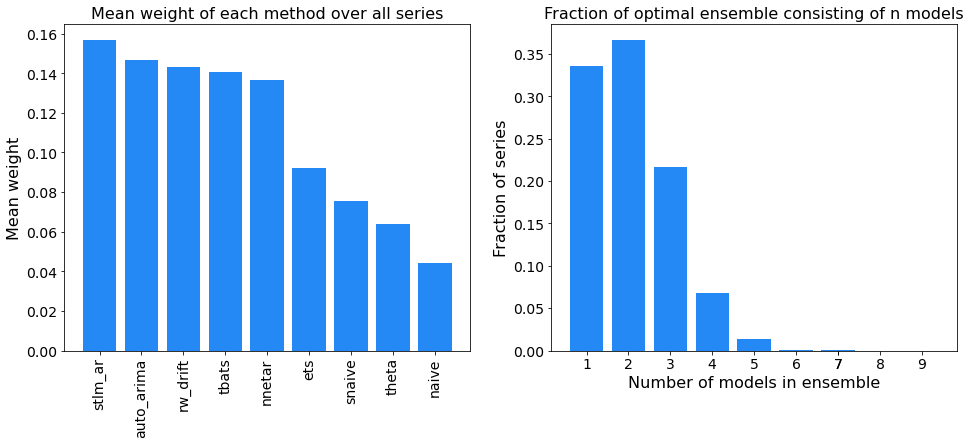}
    \caption{A display of the ex post facto best weight of the models of FFORMA on our validation set. To the left is a display of mean weights for all models. The right figure shows that 1-2 models ensemble is most common combination. 4 models our more adds up to around 7\% of total optimal solution.}
    \label{fig:opimization-9-models}
\end{figure}

\begin{figure}[H]
    \centering
    \includegraphics[width=\textwidth]{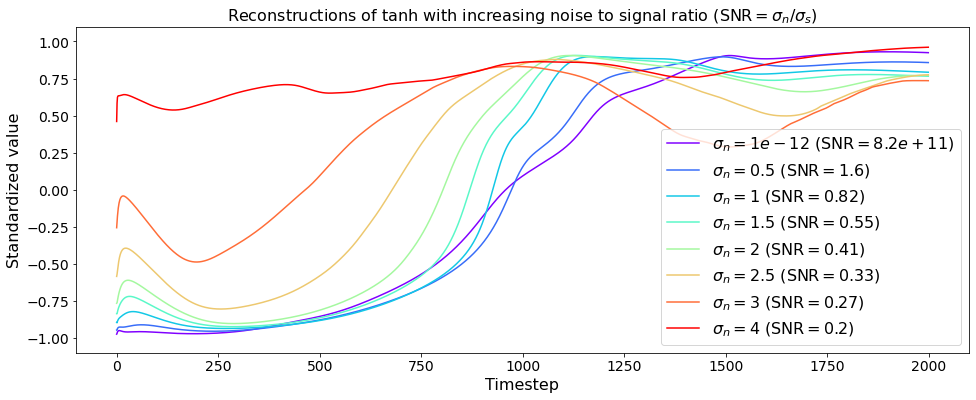}
    \caption{This figure depicts how the Autoencoder reacts to increasing noise. We generate a tanh function of 2000 data points. Then, iteratively, more and more random noise is added to it before it is given to the Autoencoder for reconstruction. The reconstruction is solid at minimal noise, shown with the purple line. The reconstruction has broken down at the other end of the spectrum, shown with the red line. This breakdown happens at around 2-3 $\sigma$ for the noise generation process or a signal-to-noise ratio of ca. 0.35.}
    \label{fig:noisy-reconstruction}
\end{figure}

\subsection{Proofs}
\label{proofs}

\begin{proof} \label{proof:model-comb}
 Given a set $\mathcal{M} \textrm{ } | \textrm{ } |\mathcal{M}|>1$ of models, $m \in \mathcal{M}$, $\mathcal{L} \in \mathcal{L}_{\Omega}$, and actual value $\mathbf{y}_{t+h}$. The ex post facto optimal combination loss will always be lower than model selection as $\mathbf{\hat{y}^*_{t+h,m}} \subseteq \mathbf{\hat{y}^{*}_{t+h,\mathcal{M}}}$ since $\mathbf{\hat{y}^*_{t+h,m}} = \mathbf{\hat{y}^{*}_{t+h,\mathcal{M}}} \iff \mathbf{\hat{y}^{*}_{t+h,\mathcal{M}}} | x_m = 1$. Thus, $\mathcal{L}(\mathbf{y}_{t+h},\mathbf{y^{*}_{t+h,\mathcal{M}}}) \leq \min\limits_{\forall m \in \mathcal{M}}(\mathcal{L}(\mathbf{y}_{t+h},\mathbf{y*_{t+h,m}}))$.
 
 Where $\min\limits_{\forall m \in \mathcal{M}}(\mathcal{L}(\mathbf{y}_{t+h},\mathbf{y^*_{t+h,m}}))$ is exactly the problem of ex post model selection.
 
 And $\mathcal{L}(\mathbf{y}_{t+h},\mathbf{y^{*}_{t+h,\mathcal{M}}})$ is the problem of ex post model combination.
 
\end{proof}

We find this interesting keeping the $T_{error}$ formula in mind and the empirical findings of several studies. \cite{M4results,makridakis2000m3,makridakis1982accuracy, makridakis1993m2} where ensembles seems to be outperforming single models, one explanation could be that the $E_{loss}(m) \geq E_{loss}(\mathcal{M}) | m \in \mathcal{M}$

\subsection{Model Hyperparameters}
\label{app:hyperparameters}

\begin{table}[ht]
    \small
    \centering
    \begin{tabular}{@{}ll@{}}
    \toprule
    \textbf{Parameter}      & \textbf{Value}           \\ \midrule
    Epochs          & 500             \\
    Batch size      & 512             \\
    Embedding dim   & 32              \\
    Hidden dim      & 128             \\
    Learning rate   & 0.002           \\
    Optimizer       & AdamW           \\
    Data-processing & Standardization \\
    Dropout         & 20\%            \\
    Weight decay    & 0.005           \\
    Max length      & 500             \\
    \bottomrule
    \end{tabular}
    \caption[LSTM Autoencoder Hyperparameters]{Hyper-parameters used to train the final LSTM autoencoder model. These hyperparameters were decided through a qualitative guessing of appropriate ranges and several days of compute devoted to a Bayesian search over the space of hyperparameters with respect to validation loss.}
    \label{tab:lstm-hyperparam}
\end{table}

\begin{table}[ht]
    \small
    \centering
    \begin{tabular}{@{}ll@{}}
    \toprule
    \textbf{Parameter}      & \textbf{Value}           \\ \midrule
    Epochs          & 12             \\
    Batch size      & 4096             \\
    Hidden dim      & 1024             \\
    Learning rate   & 0.002           \\
    Optimizer       & AdamW           \\
    Dropout         & 25.8\%            \\
    Weight decay    & 0.003064           \\
    \bottomrule
    \end{tabular}
    \caption{The hyperparameters for our final weight model, the model is a fully connected neural network.}
    \label{tab:weight-model-hyperparam}
\end{table}

%% file: Appendices/Results.tex
\section{Results}
\subsection{Comparison to FFORMA in input space}
\label{app:heatmaps}

\begin{figure}[H]
    \centering
    \includegraphics[width=\textwidth]{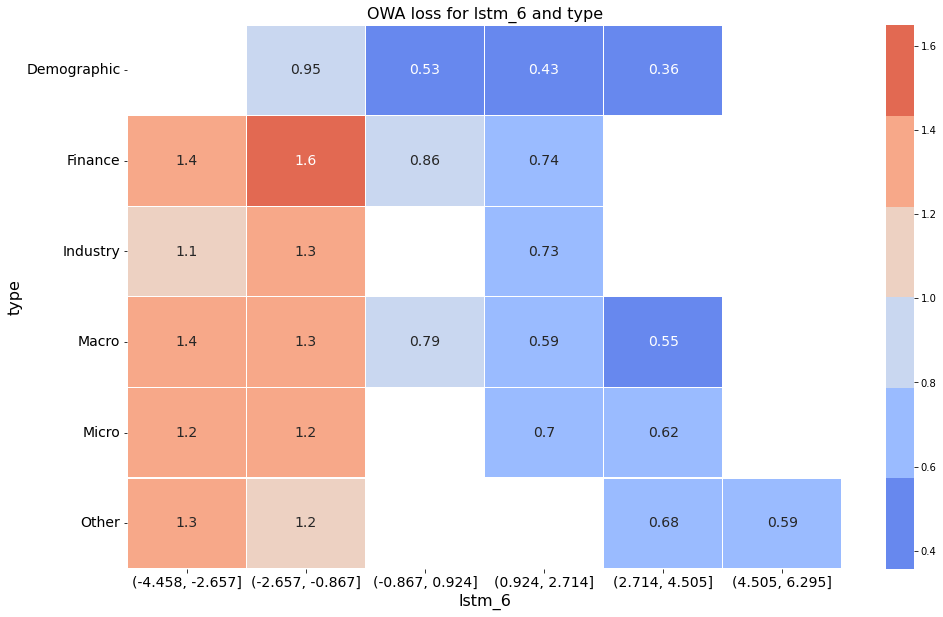}
    \caption{\textbf{$p-value<0.05$}: Shows how lstm\_6 also correlates with predictability. As lstm\_6 increases all of the series seem easier to predict. The Autoencoder have provided information which correlates with a decrease in loss}
    \label{fig:type_lstm6}
\end{figure}

\autoref{fig:type_lstm6} shows that our weighting model show a behavior when the input value of lstm\_6 is changed. Although we do not know how to interpret the machine learned feature, we can deduce that with statistical significance an increase in lstm\_6 makes our weight model more accurate.   

\begin{figure}[H]
    \centering
    \includegraphics[width=\textwidth]{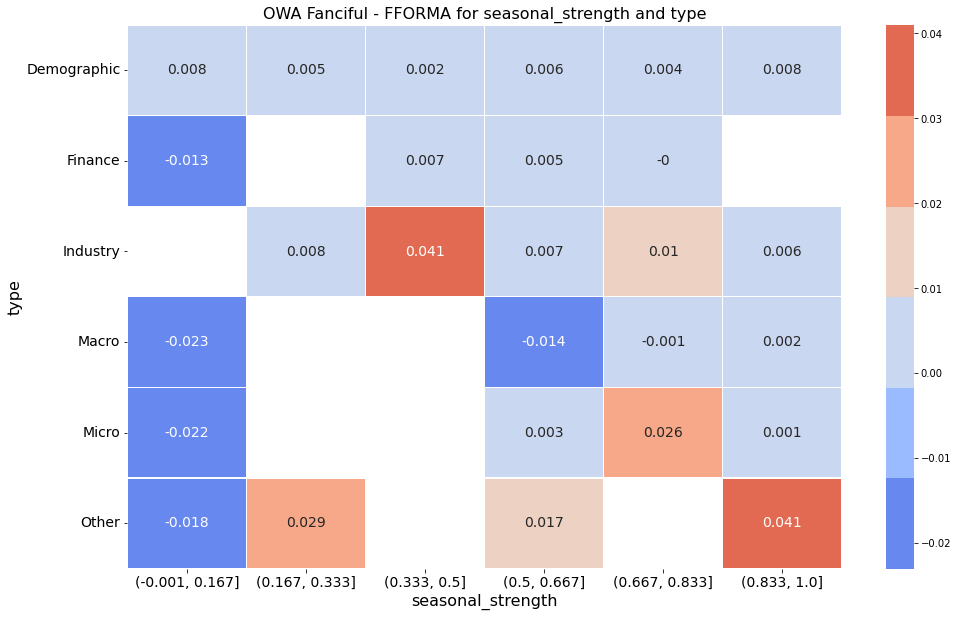}
    \caption{\textbf{$p-value<0.05$}: A strong seasonal strength correlates with a low OWA, as indicated by a strong blue color. The Demographic series seems to be most easily predicted when seasonal\_strength increases.}
    \label{fig:type_seas}
\end{figure}

\begin{figure}[H]
    \centering
    \includegraphics[width=\textwidth]{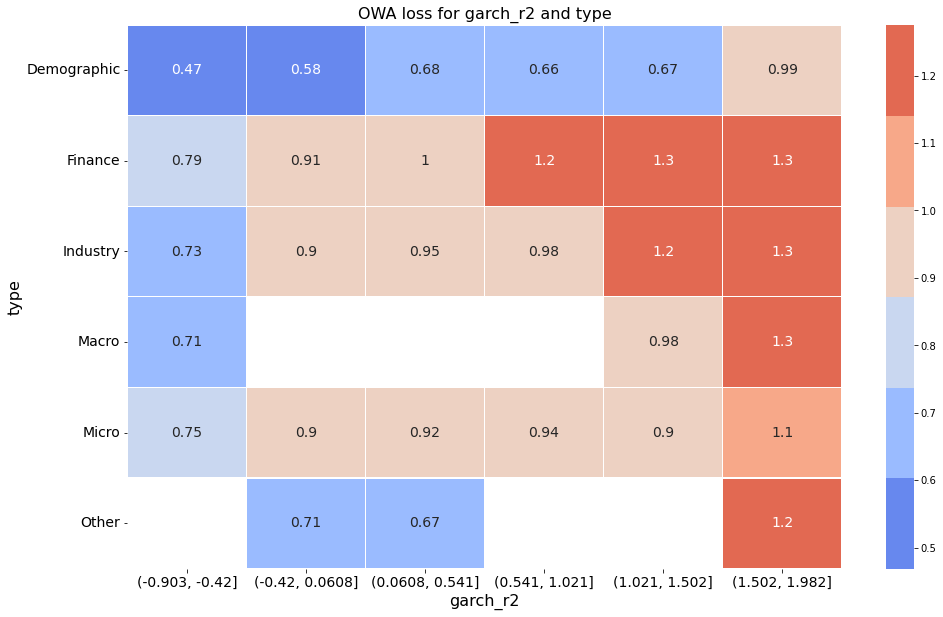}
    \caption{\textbf{$p-value<0.05$}: Garch\_r2, a measure of homoskedastic variance, also clearly correlates with a reasonable behavior from Fanciful sweep. Note how as the volatility measure increases, so does the difficulty of predicting, as indicated by a red color.}
    \label{fig:type_garchr2}
\end{figure}

\begin{figure}[H]
    \centering
    \includegraphics[width=\textwidth]{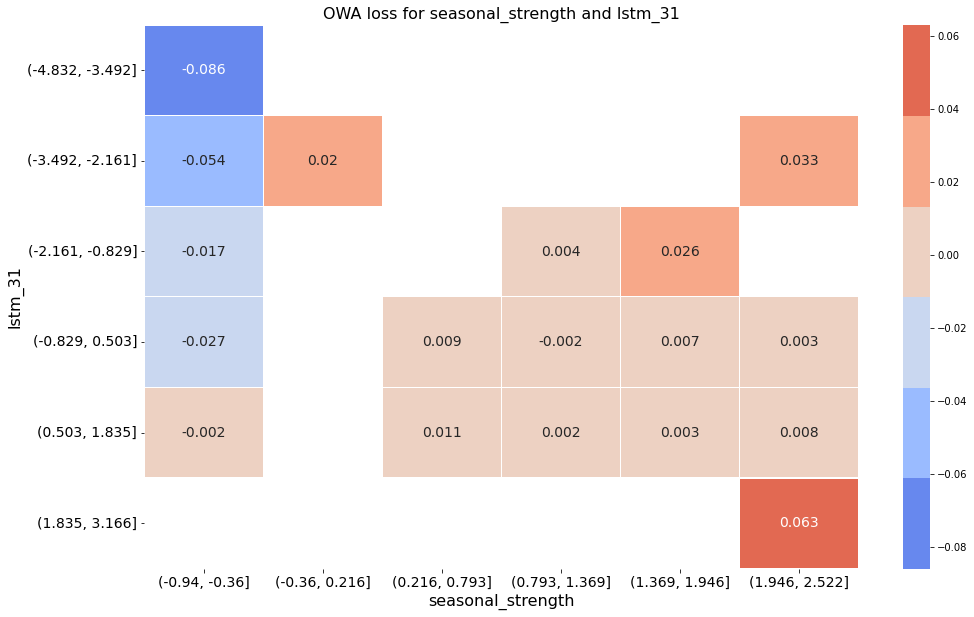}
    \caption{\textbf{$p-value<0.05$}: When plotting lstm\_31 against seasonal strength, again it seems like the auto generated feature offers extra predictive power to our net compared to FFORMA, as seasonality decreases. Note how the top right corner, has a OWA decrease of .86, with $p<0.05$}
    \label{fig:lstm31_seas_vs}
\end{figure}

\begin{figure}[H]
    \centering
    \includegraphics[width=\textwidth]{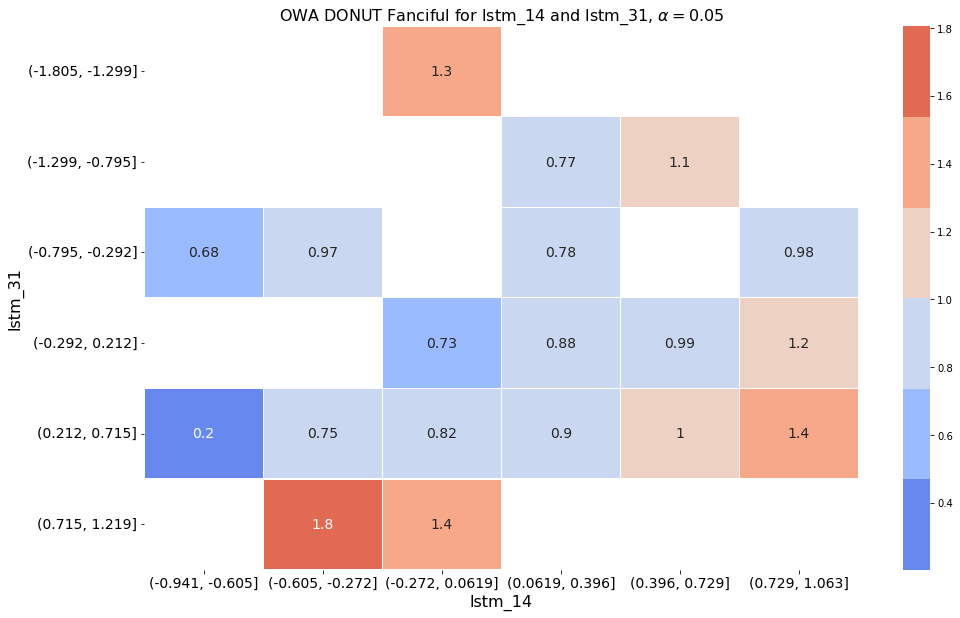}
    \caption{\textbf{$p-value<0.05$}: When both lstm\_31 and lstm\_14 take on values near their medians, the DONUT mode does better than its average by a significant. This particular combination seem to elicit a non-linear relationship between the variables.}
    \label{fig:lstm31_lstm14}
\end{figure}

It is interesting to view these results not in a void, but compared to FFORMA, to see if the LSTM features provide insight or relevant information where the statistical features do not. In \autoref{fig:lstm31_seas_vs} the values in the cells are the mean of the OWA of DONUT Fanciful less OWA FFORMA and in the relevant split of feature values. As seasonality decreases, and the value of lstm\_31 increases our weight model is increasingly outperforming FFORMA as seen by the improving blue color to the left hand side. FFORMA does not have access to this the machine learned time series feature and it indicates that lstm\_31 holds information which can be useful when seasonality is not present. Lastly \autoref{fig:lstm31_lstm14} shows a relationship between two lstm features which looks to be non-linear, this coincide with the aforementioned findings of \cite{kramer1991nonlinear} and  non-linear properties. 

\subsection{Feature Importance}
\label{app:feature-importance}

\begin{sidewaystable}
\fontsize{8}{8}\selectfont

\centering
\begin{tabular}{lllllllllll}

\toprule
{} & \textbf{lstm\_6} & \textbf{lstm\_14} & \textbf{period} & \textbf{lstm\_31} & \textbf{lstm\_13} & \textbf{lstm\_9} & \textbf{seas\_str} & \textbf{arch\_r2} & \textbf{lstm\_18} & \textbf{type} \\
\midrule
\textbf{Importance} &          4.2e-03 &           4.1e-03 &         4.1e-03 &           3.9e-03 &           3.1e-03 &          3.0e-03 &            2.9e-03 &           2.9e-03 &           2.6e-03 &       2.5e-03 \\
\textbf{t-stat    } &          2.6e+01 &           3.3e+01 &         3.0e+01 &           1.5e+01 &           1.3e+01 &          2.6e+01 &            2.2e+01 &           4.2e+01 &           2.5e+01 &       1.9e+01 \\
\textbf{p-value   } &     (1.2e-05$^c$) &      (5.3e-06$^c$) &    (7.5e-06$^c$) &      (1.3e-04$^c$) &      (1.9e-04$^c$) &     (1.3e-05$^c$) &       (2.7e-05$^c$) &      (1.9e-06$^c$) &      (1.5e-05$^c$) &  (4.8e-05$^c$) \\
\bottomrule

\\

\toprule
{} & \textbf{diff2x\_pacf5} & \textbf{garch\_r2} & \textbf{trend} & \textbf{lstm\_1} & \textbf{lstm\_26} & \textbf{seas\_acf1} & \textbf{lstm\_7} & \textbf{arch\_acf} & \textbf{trough} & \textbf{entropy} \\
\midrule
\textbf{Importance} &                2.1e-03 &            2.0e-03 &        2.0e-03 &          1.9e-03 &           1.9e-03 &             1.8e-03 &          1.8e-03 &            1.7e-03 &         1.7e-03 &          1.4e-03 \\
\textbf{t-stat    } &                2.7e+01 &            1.8e+01 &        2.5e+01 &          2.4e+01 &           2.3e+01 &             1.3e+01 &          9.2e+00 &            9.9e+00 &         2.8e+01 &          1.8e+01 \\
\textbf{p-value   } &           (1.1e-05$^c$) &       (5.5e-05$^c$) &   (1.5e-05$^c$) &     (1.7e-05$^c$) &      (2.0e-05$^c$) &        (2.2e-04$^c$) &     (7.8e-04$^c$) &       (5.9e-04$^c$) &    (1.0e-05$^c$) &     (6.1e-05$^c$) \\
\bottomrule

\\

\toprule
{} & \textbf{diff1\_acf10} & \textbf{diff1\_acf1} & \textbf{lstm\_22} & \textbf{lstm\_16} & \textbf{lstm\_23} & \textbf{lstm\_3} & \textbf{beta} & \textbf{x\_acf10} & \textbf{diff2\_acf1} & \textbf{s\_len} \\
\midrule
\textbf{Importance} &               1.4e-03 &              1.4e-03 &           1.4e-03 &           1.2e-03 &           1.2e-03 &          1.2e-03 &       1.1e-03 &           1.1e-03 &              1.1e-03 &         1.1e-03 \\
\textbf{t-stat    } &               1.0e+01 &              1.3e+01 &           2.6e+01 &           2.2e+01 &           9.5e+00 &          1.1e+01 &       1.1e+01 &           2.3e+01 &              1.1e+01 &         1.5e+01 \\
\textbf{p-value   } &          (5.0e-04$^c$) &         (1.8e-04$^c$) &      (1.3e-05$^c$) &      (2.4e-05$^c$) &      (6.8e-04$^c$) &     (4.5e-04$^c$) &  (3.5e-04$^c$) &      (2.2e-05$^c$) &         (4.5e-04$^c$) &    (1.1e-04$^c$) \\
\bottomrule

\\

\toprule
{} & \textbf{lstm\_30} & \textbf{seas\_pacf} & \textbf{lstm\_11} & \textbf{curvature} & \textbf{lstm\_28} & \textbf{lstm\_20} & \textbf{lstm\_4} & \textbf{lstm\_10} & \textbf{lstm\_8} & \textbf{hw\_beta} \\
\midrule
\textbf{Importance} &           1.1e-03 &             1.0e-03 &           1.0e-03 &            1.0e-03 &           1.0e-03 &           9.7e-04 &          9.6e-04 &           9.1e-04 &          9.0e-04 &           8.7e-04 \\
\textbf{t-stat    } &           7.9e+00 &             1.4e+01 &           1.3e+01 &            1.6e+01 &           1.3e+01 &           1.2e+01 &          1.7e+01 &           1.6e+01 &          1.2e+01 &           2.5e+01 \\
\textbf{p-value   } &       (1.4e-03$^b$) &        (1.7e-04$^c$) &      (2.0e-04$^c$) &       (9.6e-05$^c$) &      (2.0e-04$^c$) &      (2.4e-04$^c$) &     (7.8e-05$^c$) &      (8.0e-05$^c$) &     (2.4e-04$^c$) &      (1.5e-05$^c$) \\
\bottomrule

\\

\toprule
{} & \textbf{hw\_gamma} & \textbf{e\_acf10} & \textbf{hurst} & \textbf{lstm\_29} & \textbf{linearity} & \textbf{lstm\_0} & \textbf{e\_acf1} & \textbf{lstm\_17} & \textbf{flat\_spots} & \textbf{lstm\_2} \\
\midrule
\textbf{Importance} &            8.6e-04 &           8.6e-04 &        8.6e-04 &           8.4e-04 &            8.3e-04 &          8.2e-04 &          7.9e-04 &           7.5e-04 &              7.1e-04 &          7.0e-04 \\
\textbf{t-stat    } &            9.1e+00 &           6.8e+00 &        9.5e+00 &           1.3e+01 &            1.4e+01 &          1.7e+01 &          5.7e+00 &           1.5e+01 &              1.8e+01 &          7.8e+00 \\
\textbf{p-value   } &       (8.0e-04$^c$) &       (2.4e-03$^b$) &   (6.8e-04$^c$) &      (1.8e-04$^c$) &       (1.7e-04$^c$) &     (7.2e-05$^c$) &      (4.6e-03$^b$) &      (1.0e-04$^c$) &         (5.1e-05$^c$) &      (1.4e-03$^b$) \\
\bottomrule

\\

\toprule
{} & \textbf{lstm\_27} & \textbf{nperiods} & \textbf{stability} & \textbf{nonlinearity} & \textbf{lstm\_21} & \textbf{hw\_alpha} & \textbf{unitroot\_kpss} & \textbf{garch\_acf} & \textbf{ARCH.LM} & \textbf{diff1x\_pacf5} \\
\midrule
\textbf{Importance} &           7.0e-04 &           7.0e-04 &            7.0e-04 &               6.9e-04 &           6.6e-04 &            6.5e-04 &                 6.4e-04 &             5.9e-04 &          5.8e-04 &                5.7e-04 \\
\textbf{t-stat    } &           1.3e+01 &           7.6e+00 &            1.3e+01 &               1.0e+01 &           9.9e+00 &            9.5e+00 &                 1.5e+01 &             1.5e+01 &          1.3e+01 &                9.8e+00 \\
\textbf{p-value   } &      (1.8e-04$^c$) &       (1.6e-03$^b$) &       (2.3e-04$^c$) &          (5.4e-04$^c$) &      (5.9e-04$^c$) &       (6.8e-04$^c$) &            (1.2e-04$^c$) &        (1.2e-04$^c$) &     (2.2e-04$^c$) &           (6.1e-04$^c$) \\
\bottomrule

\\

\toprule
{} & \textbf{x\_acf1} & \textbf{lstm\_19} & \textbf{unitroot\_pp} & \textbf{lstm\_12} & \textbf{alpha} & \textbf{diff2\_acf10} & \textbf{lstm\_5} & \textbf{seas\_per} & \textbf{x\_pacf5} & \textbf{lstm\_24} \\
\midrule
\textbf{Importance} &          5.6e-04 &           4.9e-04 &               4.7e-04 &           4.7e-04 &        4.6e-04 &               4.5e-04 &          4.4e-04 &            3.7e-04 &           3.7e-04 &           3.6e-04 \\
\textbf{t-stat    } &          3.2e+00 &           4.0e+00 &               7.8e+00 &           2.0e+01 &        5.4e+00 &               1.2e+01 &          9.5e+00 &            7.3e+00 &           4.8e+00 &           7.2e+00 \\
\textbf{p-value   } &       (3.2e-02$^a$) &        (1.6e-02$^a$) &           (1.5e-03$^b$) &      (3.4e-05$^c$) &    (5.5e-03$^b$) &          (3.0e-04$^c$) &     (7.0e-04$^c$) &        (1.9e-03$^b$) &       (8.7e-03$^b$) &       (2.0e-03$^b$) \\
\bottomrule

\\

\toprule
{} & \textbf{lstm\_25} & \textbf{cross\_ps} & \textbf{peak} & \textbf{spike} & \textbf{lstm\_15} & \textbf{lumpiness} \\
\midrule
\textbf{Importance} &           3.3e-04 &            3.2e-04 &       2.5e-04 &        2.4e-04 &           1.5e-04 &            1.3e-04 \\
\textbf{t-stat    } &           6.8e+00 &            8.1e+00 &       4.9e+00 &        4.0e+00 &           2.9e+00 &            4.0e+00 \\
\textbf{p-value   } &       (2.5e-03$^b$) &        (1.3e-03$^b$) &   (8.1e-03$^b$) &     (1.7e-02$^a$) &        (4.2e-02$^a$) &         (1.6e-02$^a$) \\
\bottomrule

\end{tabular}

\caption{Table summarizing the feature importances found for the trained weighting network with corresponding $t$-statistics and $p$-values. Significance levels: $^a$$\alpha=0.05$, $^b$$\alpha=0.01$, and $^c$$\alpha=0.001$.}
\label{tab:feature-importance}

\end{sidewaystable}